\documentclass{article}

\usepackage[preprint]{neurips_2026}      % arXiv / preprint

\usepackage[utf8]{inputenc}
\usepackage[T1]{fontenc}
\usepackage{hyperref}
\usepackage{url}
\usepackage{booktabs}
\usepackage{amsfonts}
\usepackage{amsmath}
\usepackage{amssymb}
\usepackage{nicefrac}
\usepackage{microtype}
\usepackage{graphicx}
\usepackage{tikz}
\usetikzlibrary{arrows.meta,fit,positioning}
\usepackage{cleveref}
\usepackage{algorithm}
\usepackage{algpseudocode}

\usepackage{xcolor}
\usepackage{etoolbox}
\usepackage{marginnote}

\newif\ifshownotes
\shownotestrue

\definecolor{elliotcolor}{RGB}{0, 102, 204}    % blue
\definecolor{chiraagcolor}{RGB}{204, 85, 0}    % burnt orange
\definecolor{sidcolor}{RGB}{102, 51, 153}      % purple

\title{Pre-trained Tabular Foundation Models as Versatile Summary Networks for Neural Posterior Estimation}

\author{%
  Elliot Pickens\textsuperscript{1}, Chiraag Gohel\textsuperscript{2}, Sidharth Satya\textsuperscript{3} \\
  \textsuperscript{1}University of Pennsylvania, \textsuperscript{2}The George Washington University, \textsuperscript{3}Brown University \\
  \textsuperscript{1}\texttt{epickens@seas.upenn.edu}, \textsuperscript{2}\texttt{chiraaggohel@gwu.edu}, \textsuperscript{3}\texttt{sidharth\_satya@brown.edu} \\
}

\begin{document}

\maketitle

% !TEX root = ../main.tex
\begin{abstract}

% option 2.5 
In this work, we study TabPFN as a training-free, modular summary network for simulation-based Bayesian inference (SBI). Tabular foundation models such as TabPFN are pretrained on broad families of synthetic tabular data-generating processes and adapt at test time through in-context learning, making them natural candidates for SBI, where posterior estimation often depends on learning informative summaries of simulated observations. We propose PFN-NPE: a general recipe that uses a pretrained TabPFN encoder as a fixed summary network for simulator outputs, then pairs the resulting summaries with a downstream inference head chosen for the problem. With normalizing flows as the default inference head, PFN-NPE matches established posterior approximation methods and sometimes outperforms them. More importantly, diagnostic probes show that the TabPFN-derived summaries often preserve useful posterior location and marginal information. These analyses also reveal a limitation in that TabPFN-derived summaries may struggle to represent the joint posterior structure even when the marginals are well recovered. Still, our experiments show that TabPFN can serve as an effective summary network across a diverse set of SBI settings, with the inference network left modular and task-dependent.

\end{abstract}

\renewcommand{\thefootnote}{}
\footnotetext{Research code available at \href{https://github.com/epickens/pfn-npe}{\texttt{epickens/pfn-npe}}}
% !TEX root = ../main.tex
\section{Introduction}
\label{sec:introduction}

Many scientific models use a simulator \(x \sim p(x \mid \theta)\) whose
likelihood is intractable or expensive to evaluate. Simulation-based inference
(SBI) handles this setting by training conditional posterior, likelihood, or
ratio estimators on simulated parameter--observation pairs
\citep{cranmer2020frontier,papamakarios2016fast,greenberg2019automatic}. Once
trained, these estimators can answer posterior queries for a fixed observation
\(x_o\), but their quality depends on the simulation budget and on how
simulator observations are represented.

Prior-fitted networks (PFNs) offer another route to amortization. A PFN is
trained offline on synthetic datasets drawn from a prior over data-generating
processes and adapts at test time through in-context examples
\citep{muller2022transformers}. TabPFN applies this idea to tabular prediction
without task-specific gradient updates
\citep{hollmann2023tabpfn,hollmann2025tabpfn}. Because simulator pairs
\(\{(\theta_i,x_i)\}_{i=1}^n\) can be written as tabular contexts, TabPFN is a
natural candidate for SBI. Recent work uses this connection in NPE-PFN, which
queries TabPFN directly as an autoregressive posterior sampler
\citep{vetter2025npepfn}.

We study a different use of TabPFN: a frozen summary model for neural posterior
estimation. PFN-NPE builds a parameter-wise TabPFN context, extracts frozen
embeddings for simulated and observed outputs, applies a fixed projection, and
trains a normalizing flow for \(q_\phi(\theta \mid s(x))\). This shifts
posterior sampling to conditional density estimation over fixed TabPFN-derived
summaries. It also lets us test which posterior features those summaries
preserve.

We make three contributions. First, we formulate PFN-NPE as a frozen-summary
neural posterior estimation method. Second, we evaluate it across standard
SBIBM benchmarks, controlled nuisance-coordinate variants, and structured
time-series simulators under matched simulation budgets
\citep{lueckmann2021benchmarking}. Third, we use marginal C2ST, rank-space
C2ST, and quantile probes to measure posterior discrepancies beyond aggregate
joint C2ST \citep{lopezpaz2017revisiting}.

The empirical comparison shows that performance depends on the task. With a
budget of \(10{,}000\) simulations, NPE-PFN has the lowest mean joint C2ST on
most reported tasks. PFN-NPE performs best on mixture and nuisance-coordinate
settings and improves over the learned-summary NPE baseline on most tasks
where both methods are available. Diagnostics show that TabPFN-derived
summaries often encode one-dimensional posterior location, scale, and quantiles
in a linearly accessible form. But joint and rank-space C2ST remain elevated on
high-gap tasks, so accurate multivariate posterior structure remains the main
challenge for this frozen-summary method.

% !TEX root = ../main.tex
\section{Background and Related Work}
\label{sec:background}

\subsection{Simulation-based inference}
\label{sec:background:sbi}

Simulation-based inference (SBI) addresses Bayesian inference problems where
simulator samples \(x \sim p(x \mid \theta)\) are available but likelihood
evaluation is inaccurate, expensive, or unavailable. Modern neural SBI trains
reusable posterior, likelihood, or ratio estimators from simulated
parameter--observation pairs \citep{cranmer2020frontier}. Neural posterior
estimation fits \(q_\phi(\theta \mid x)\) directly, often with normalizing
flows \citep{papamakarios2016fast,greenberg2019automatic,durkan2019neural}.
Neural likelihood estimation fits \(q_\phi(x \mid \theta)\) and combines it
with the prior for posterior sampling \citep{papamakarios2019sequential}.
Neural ratio estimation learns likelihood-ratio functions through
classification objectives
\citep{hermans2020likelihoodfree,durkan2020contrastive}. Flow-matching methods
provide another route to conditional posterior sampling
\citep{wildberger2023fmpe}. Recent amortized Bayesian inference work also uses
transformers and transformer-backed diffusion models
\citep{mittal2025amortized,gloeckler2024allinone}.

SBI needs diagnostics on posterior samples. SBIBM provides standardized
simulator tasks, reference posterior samples, and benchmark metrics
\citep{lueckmann2021benchmarking}. We use classifier two-sample tests (C2ST),
where chance-level accuracy means that approximate and reference posterior
samples are hard to distinguish \citep{lopezpaz2017revisiting}. Aggregate C2ST
scores summarize sample quality but do not show whether errors come from
marginal calibration, dependence structure, or coverage
\citep{talts2018sbc,hermans2022crisis,lemos2023sampling}. This motivates the
marginal, rank-space, and quantile diagnostics used in \cref{sec:diagnostic}.

\subsection{Prior-fitted networks}
\label{sec:background:pfn}

Prior-data fitted networks (PFNs) amortize inference over a distribution of
tasks. A PFN is trained offline on synthetic datasets sampled from a prior over
data-generating processes. At test time, it receives a dataset in context and
returns predictions in a single forward pass \citep{muller2022transformers}.
This connects Bayesian prediction, meta-learning, and in-context learning.

TabPFN applies this framework to tabular prediction. The original model was
trained on synthetic tabular classification tasks and predicts labels for
small datasets without task-specific gradient updates
\citep{hollmann2023tabpfn}. Recent TabPFN variants extend this setting to
broader tabular tasks while preserving fast in-context prediction
\citep{hollmann2025tabpfn}. In SBI, simulated pairs \((\theta_i,x_i)\) can be
formatted as tabular contexts, with simulator parameters as prediction targets.
This makes TabPFN a reusable component to test for posterior inference.

The key question is which posterior features this supervised tabular
pretraining preserves. TabPFN is trained under its pretraining prior, so its
behavior depends on the match between that prior, the in-context encoding, and
the simulator family at test time. Theoretical work on PFNs shows that
architectural symmetries alone do not guarantee Bayesian consistency under
arbitrary task shifts \citep{nagler2023statistical}. For SBI, this shifts
attention to empirical diagnostics of the representation, posterior samples,
and task-dependent failures.

\subsection{TabPFN for SBI}
\label{sec:background:tabpfn_sbi}

The closest prior work is NPE-PFN \citep{vetter2025npepfn}, which uses TabPFN
as a training-free posterior estimator. Given simulated
parameter--observation pairs, NPE-PFN formats each parameter coordinate as a
scalar regression target and queries TabPFN autoregressively to sample from the
posterior. This gives a fast in-context SBI method with strong simulation
efficiency in several benchmark settings. Its samples are constrained by
TabPFN's context size, scalar target interface, and autoregressive parameter
ordering.

PFN-NPE uses TabPFN through a different interface. It freezes TabPFN, extracts
intermediate embeddings as simulator-output summaries, and trains a
conditional density estimator for \(\theta \mid s(x)\). This separates the
foundation-model representation from posterior sampling. It also lets us test
whether frozen TabPFN summaries carry posterior information that a standard
neural posterior estimator can use. \Cref{sec:method} formalizes this
construction.

% !TEX root = ../main.tex
\section{Method}
\label{sec:method}

\subsection{Problem setup}
\label{sec:method:setup}

For each simulator task, we construct pairs
\(\mathcal{D}=\{(\theta_i,x_i)\}_{i=1}^n\), where
\(\theta_i \sim p(\theta)\) and \(x_i \sim p(x \mid \theta_i)\). Given a new
observation \(x_o\), we target the posterior \(p(\theta \mid x_o)\). We split
the simulated pairs into training and validation sets. We use the training set
to define the TabPFN contexts and fit posterior estimators. The validation set
supports early stopping and hyperparameter selection where applicable. Unless
otherwise stated, experiments use \(n=10{,}000\) training simulations, \(2{,}000\)
validation simulations, and the ten reference observations provided by
\texttt{SBIBM} for each task \citep{sbibm-pmlr-v130-lueckmann21a}.

\subsection{PFN-NPE}
\label{sec:method:pfn_npe}

PFN-NPE uses TabPFN as a frozen summary network for simulator outputs, then
passes the summaries to a posterior estimator. The procedure has three
configurable components: the in-context TabPFN construction, the embedding
transformation, and the conditional density estimator. In our experiments, the
TabPFN context is a uniformly sampled subset of \(m=\min(1000,n)\) training
simulations. The summary network is the TabPFN v2 regressor with a single
ensemble member \citep{hollmann2023tabpfn,hollmann2025tabpfn}. The TabPFN
weights remain fixed throughout.

\begin{figure}
    \centering
    \includegraphics[width=0.95\linewidth]{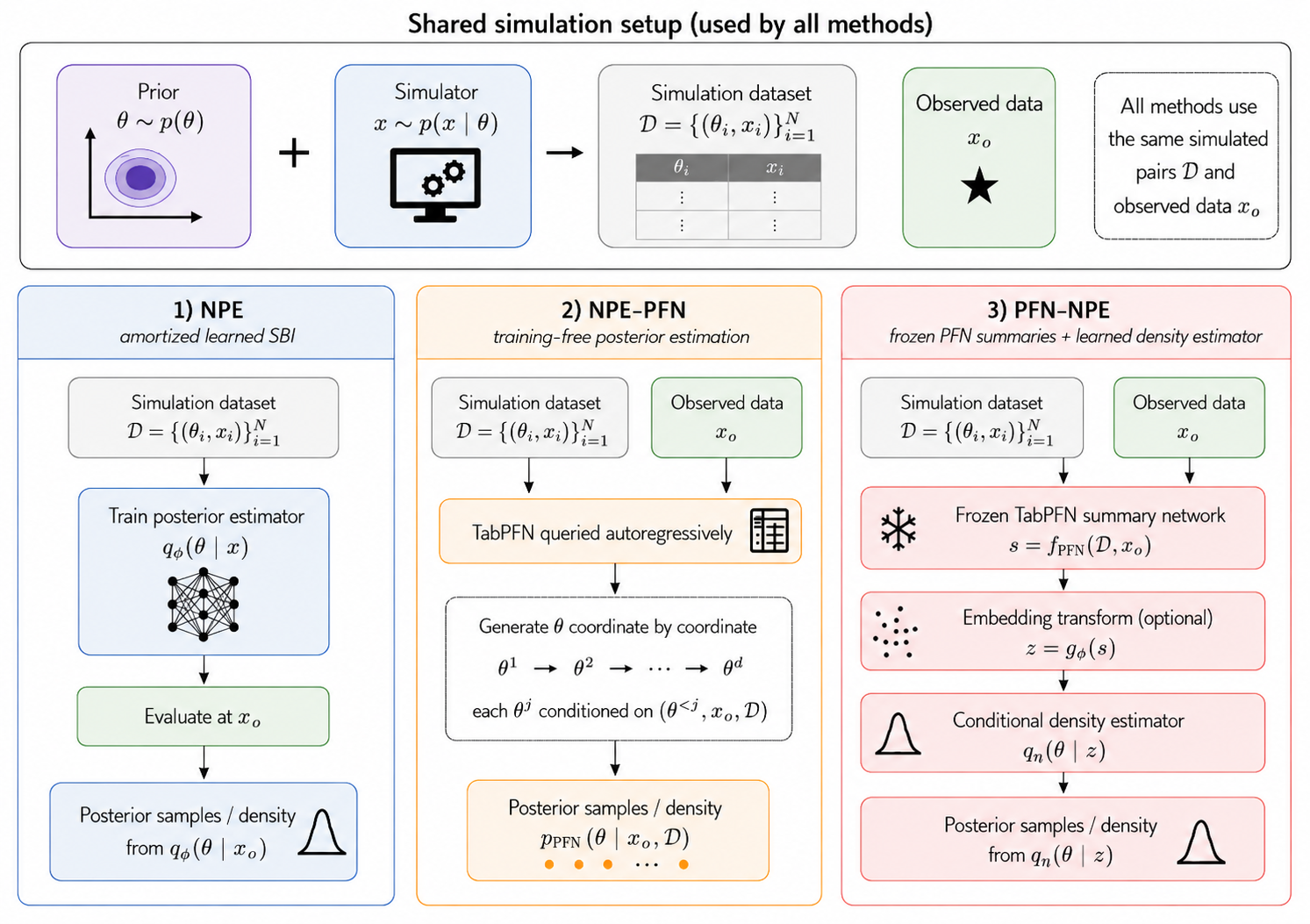}
    \caption{Three posterior-inference workflows using the same simulated
        training set $\mathcal{D}=\{(\theta_i,x_i)\}_{i=1}^N$ and observed data
        $x_o$. Standard NPE trains an amortized posterior estimator directly on
        simulations. NPE-PFN uses a pretrained TabPFN autoregressively at
        inference time to draw posterior samples. PFN-NPE uses a frozen TabPFN
        as a summary network and trains a conditional density estimator on the
        resulting embeddings.}
    \label{fig:pfn_npe_architecture}
\end{figure}

The summary is parameter-wise. PFN-NPE first samples a shared context index set
\(I_{\mathrm{ctx}}\) of size \(m\). For each parameter coordinate
\(d \in \{1,\ldots,D_\theta\}\), it forms a scalar TabPFN regression context
\[
  \mathcal{C}_d
  =
  \{(x_i,\theta_{i,d}) : i \in I_{\mathrm{ctx}}\},
\]
where simulator outputs \(x_i\) are tabular features and the scalar parameter
coordinate \(\theta_{i,d}\) is the regression target. Every coordinate uses the
same context observations; only the scalar target changes. With
\(\mathcal{C}_d\) fixed, we query TabPFN on each training simulation,
validation simulation, and reference observation. Let \(h_d(x)\) denote the
final-layer query embedding for output \(x\) under coordinate context
\(\mathcal{C}_d\).

PFN-NPE repeats this embedding pass once per parameter coordinate and
concatenates the embeddings,
\[
  h_{\mathrm{raw}}(x)
  =
  \big[h_1(x),\ldots,h_{D_\theta}(x)\big].
\]
This gives a fixed representation of simulator output \(x\). By default, we
fit PCA on \(\{h_{\mathrm{raw}}(x_i)\}_{i=1}^{n}\) from the training
simulations and keep the projection fixed. The same projection maps training,
validation, and reference summaries to
\[
  s(x)=g\!\left(h_{\mathrm{raw}}(x)\right)\in\mathbb{R}^{64}.
\]
We use PCA for \(g\) by default. \Cref{app:pca_no_pca} reports a matched
PCA/no-PCA ablation.

All TabPFN queries happen during summary precomputation. The conditional
density estimator then trains on fixed pairs
\(\{(s(x_i),\theta_i)\}_{i=1}^{n}\), uses validation summaries for early
stopping, and draws samples for a reference observation from the trained model
at \(s(x_o)\). This makes the frozen summary stage reusable across posterior
estimators and limits task-specific training to the conditional density
estimator.

\subsection{Conditional density estimator}
\label{sec:method:density_estimator}

PFN-NPE maps each simulator output \(x_i\) to a fixed summary \(s(x_i)\). The
posterior estimator trains on pairs \(\{(s(x_i),\theta_i)\}_{i=1}^{n}\) and
represents \(q_\phi(\theta \mid s(x))\). TabPFN is queried only during summary
precomputation, so training, validation, and reference observations all use the
same summary map.

All experiments use a neural spline flow (NSF) for \(q_\phi\)
\citep{durkan2019neural}. We standardize parameter vectors with the
training-set mean \(\mu_\theta\) and standard deviation \(\sigma_\theta\), then
train the flow by minimizing negative log likelihood,
\[
  \mathcal{L}(\phi)
  =
  -\frac{1}{n}\sum_{i=1}^{n}
  \log q_\phi\!\left(
    \frac{\theta_i-\mu_\theta}{\sigma_\theta}
    \,\middle|\, s(x_i)
  \right).
\]
Validation simulations determine early stopping and hyperparameter selection.
\Cref{app:training-details} reports architecture and optimization details.

At inference time, we embed a reference observation \(x_o\) with the same
TabPFN contexts \(\{\mathcal{C}_d\}_{d=1}^{D_\theta}\), transform it with the
fixed map \(g\), and pass it to the trained flow as \(s(x_o)\). We draw
posterior samples from \(q_\phi(\cdot \mid s(x_o))\) in standardized parameter
space and transform them back to the original scale. In the local benchmark,
each method draws \(1000\) posterior samples per reference observation.

The learned-summary NPE baseline uses the same NSF posterior estimator and
training protocol with a task-specific MLP summary map
\(a_\eta(x)\in\mathbb{R}^{64}\). The MLP and NSF train jointly on the same
simulations, validation split, seeds, and reference observations. This gives a
matched learned-summary comparison at the same summary dimension.

% !TEX root = ../main.tex
\section{Empirical Comparison}
\label{sec:experiments}

We evaluate PFN-NPE across standard SBI benchmarks, nuisance-rich observation
variants, and structured time-series simulators. The comparison includes direct
NPE-PFN inference, a learned-summary NPE control with the same NSF posterior
estimator, and external neural SBI baselines where available. We ask how
TabPFN-derived summaries perform under matched simulation budgets and where
posterior-sample discrepancies appear in joint, marginal, and rank-space
diagnostics. We first define the task suite and metrics, then report
budget-scaling results and the fixed-budget pattern that motivates
\cref{sec:diagnostic}.

\subsection{Tasks and metrics}
\label{sec:experiments:tasks}

The task suite covers common SBI regimes: standardized benchmark simulators,
nuisance-rich observations, and structured time series. The SBIBM suite
provides standardized simulators, fixed reference observations, reference
posterior samples, and established neural SBI baselines
\citep{lueckmann2021benchmarking}. It covers low-dimensional multimodal
tasks, Gaussian models with analytic structure, high-dimensional parameter
tasks, and mechanistic ODE simulators. Controlled distractor variants append
independent Gaussian-mixture nuisance coordinates to simulator outputs and
permute the columns. Since these nuisance coordinates are independent of
$\theta$, the base-task reference posterior remains valid as the observation
dimension grows. The custom time-series tasks include an AR(1) Gaussian
process, an Ornstein--Uhlenbeck process, and a solar-dynamo simulator, giving
structured sequential observations with dimensions between 50 and 100.

Each default run uses $10{,}000$ training simulations, $2{,}000$ validation
simulations, and the ten fixed reference observations for the task. Each
method returns $1000$ posterior samples per reference observation. We evaluate
three random seeds, average metrics over reference observations within
each seed, and report means and standard deviations across seeds.
Budget-scaling figures vary the number of training simulations and keep the
reference observations, posterior sample count, and evaluation protocol fixed.

We use joint classifier two-sample test (C2ST) as the main posterior-quality
metric \citep{lopezpaz2017revisiting}. C2ST trains a classifier to distinguish
approximate posterior samples from reference posterior samples. Accuracy
$0.5$ means the sample sets are indistinguishable, while larger values indicate
greater discrepancy. The joint score applies this test in multivariate
\(\theta\)-space.

We also report marginal and rank-space C2ST to localize posterior
discrepancies. The marginal score averages one-dimensional C2ST values over
parameter coordinates,
\[
  \mathrm{C2ST}_{\mathrm{marg}}
  =
  \frac{1}{D_\theta}
  \sum_{d=1}^{D_\theta}
  \mathrm{C2ST}\!\left(q(\theta_d \mid x_o),
  p(\theta_d \mid x_o)\right).
\]
This score summarizes one-dimensional marginal mismatch. The rank-space score
maps approximate and reference samples through a pooled empirical CDF in each
parameter coordinate, then applies C2ST to the transformed samples. This
transform equalizes one-dimensional marginals while preserving sensitivity to
dependence structure and other joint-distribution errors. These diagnostics
identify cases where PFN-derived summaries preserve marginal posterior
information but leave joint discrepancies.

For quantile diagnostics, we use pinball loss at
$\tau \in \{0.05,0.25,0.5,0.75,0.95\}$,
\[
  \ell_\tau(y, \hat q_\tau)
  =
  \max\!\left\{\tau (y-\hat q_\tau),
  (\tau-1)(y-\hat q_\tau)\right\}.
\]
These scores give another view of marginal posterior calibration and
one-dimensional uncertainty.

\subsection{Budget-scaling comparison}
\label{sec:experiments:sbibm}

\begin{table}[t]
  \centering
  \caption{Compact $10{,}000$-simulation summary by task family. Counts report
  how often each local method has the lowest mean joint C2ST within the family.
  The median columns summarize PFN-NPE and the best available local method per
  task. Lower C2ST is better.}
  \label{tab:compact_family_summary}
  \resizebox{\linewidth}{!}{% Auto-generated by scripts/paper_comparison_tables.py. Do not edit by hand.
\begin{tabular}{lrrrrcc}
  \toprule
  Task family & \# tasks & PFN-NPE best & Learned summary best & NPE-PFN best & Median PFN-NPE & Median best \\
  \midrule
  SBIBM reference & 10 & 1 & 1 & 8 & $0.641$ & $0.577$ \\
  Distractor variants & 4 & 2 & 0 & 2 & $0.628$ & $0.576$ \\
  Time series & 3 & 1 & 1 & 1 & $0.866$ & $0.857$ \\
  \bottomrule
\end{tabular}
}
\end{table}

\begin{figure}[t]
  \centering
  \includegraphics[width=\linewidth]{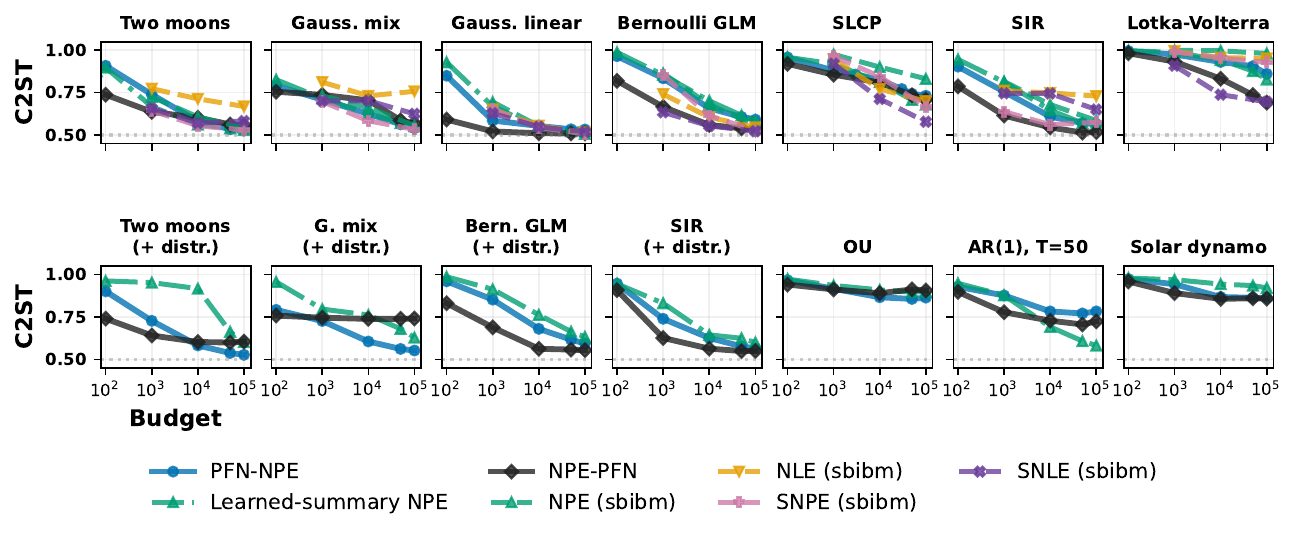}
  \vspace{-2em}
  \caption{Joint C2ST vs.\ simulation budget across the standard and extended task suite. The first row shows the seven standard SBIBM tasks. The second row shows four distractor variants followed by three time-series tasks. The dotted line marks the ideal C2ST value of $0.5$.}
  \label{fig:budget_scaling_standard_plus_extended}
\end{figure}

\Cref{fig:budget_scaling_standard_plus_extended} summarizes joint C2ST across
the standard and extended task suite. The standard SBIBM row shows strong task
dependence. The Gaussian linear tasks approach the C2ST floor for all local
methods by \(10^4\)--\(10^5\) simulations. Two moons and Gaussian mixture
improve steadily with budget, with PFN-NPE close to the learned-summary NPE
baseline at the default budget. Bernoulli GLM and SIR favor autoregressive
NPE-PFN through the default-budget region.

SLCP and Lotka--Volterra define the high-error end of the standard suite.
PFN-based curves retain high joint C2ST across the displayed budget range on
these tasks. Several external baseline curves decrease at larger budgets,
especially shipped SNLE on SLCP, showing that the high-error regime depends on
the inference interface. This pattern motivates the diagnostic analysis:
additional simulations help on several tasks, while structured joint posterior
geometry remains limiting for PFN-based approaches.

The extended row tests nuisance-coordinate robustness and structured
time-series behavior. In the distractor variants, PFN-NPE has its clearest
advantage on \texttt{two\_moons\_distractors} and
\texttt{gaussian\_mixture\_distractors}, reaching \(0.582\pm0.02\) and
\(0.605\pm0.04\) at \(10^4\) simulations. Bernoulli GLM and SIR distractor
variants follow their base-task pattern and favor NPE-PFN. The
time-series tasks retain high C2ST across local methods, with different
default-budget leaders across OU, AR(1), and solar dynamo. Overall, PFN-NPE
gains concentrate in mixture and nuisance-coordinate settings, while sequence
tasks leave substantial joint posterior error across local methods.

\subsection{Fixed-budget summary}
\label{sec:experiments:summary}

At the default budget of $10{,}000$ training simulations, the joint C2ST table
shows a task-dependent ranking across the three local methods
(\cref{tab:c2st_joint_full}). NPE-PFN attains the lowest mean joint C2ST on
11 of 17 tasks, with its strongest relative performance on Bernoulli GLM
variants, SIR variants, and Lotka--Volterra. Direct autoregressive TabPFN
inference is therefore a strong fixed-budget baseline on many standard
benchmark tasks.

PFN-NPE gives the lowest joint C2ST on \texttt{gaussian\_mixture},
\texttt{gaussian\_mixture\_distractors}, \texttt{two\_moons\_distractors},
and \texttt{ou}, and it improves over the learned-summary NPE baseline on most
tasks where that baseline is available. The high-C2ST rows in
\cref{tab:c2st_joint_full} show where all local methods struggle, including
SLCP, Lotka--Volterra, OU, solar dynamo, and AR(1). These results set up the
diagnostic analysis in \cref{sec:diagnostic}: PFN-NPE is strongest in mixture
and nuisance-coordinate regimes, while structured joint posterior geometry
remains the main failure mode.

% !TEX root = ../main.tex
\section{Diagnostic Analysis of Marginals and Joints}
\label{sec:diagnostic}

The empirical results in \cref{sec:experiments} show task-dependent behavior
across PFN-NPE, NPE-PFN, and learned-summary NPE. We use probes to ask two
questions: whether frozen TabPFN summaries carry marginal posterior
information, and whether PFN-NPE samples match the joint posterior after the
conditional density estimator is fitted.

A linear probe keeps the representation fixed and trains a simple linear
predictor for a chosen target
\citep{alain2017understanding,hewitt2019structural}. High validation
performance means the target information is linearly accessible. Low validation
performance does not rule out nonlinear or task-dependent encodings. Prior TabPFN
interpretability work uses similar probes to study linearly decodable
structure inside TabPFN representations
\citep{gupta2026lookingglass}.

This matters for PFN-NPE because the summary has built-in coordinate
structure. For each parameter coordinate \(d\), PFN-NPE builds a TabPFN
regression context with \(\theta_d\) as the target and stores the query
embedding \(h_d(x)\). The raw summary concatenates these chunks as
\([h_1(x),\ldots,h_{D_\theta}(x)]\). The probes test what each frozen chunk
makes linearly accessible. The C2ST diagnostics then evaluate samples after PCA
and the normalizing flow.

\subsection{Linear probes for marginal information}
\label{sec:diagnostic:moments}

\begin{figure}[t]
  \centering
  \includegraphics[width=\linewidth]{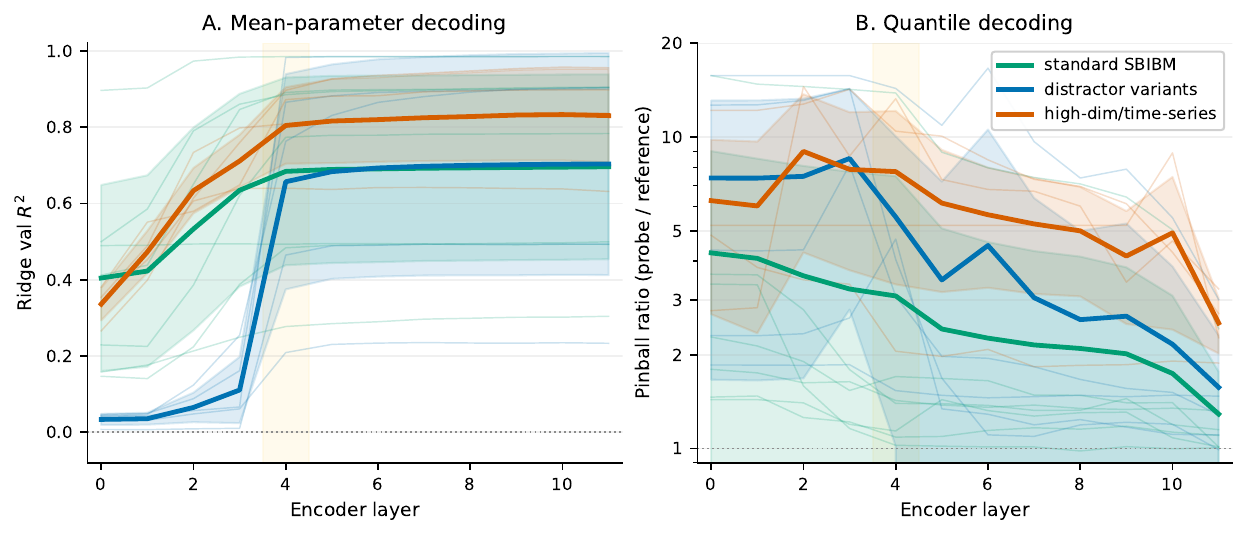}
  \caption{Layer-wise linear-probe diagnostics for parameter information in
  frozen TabPFN summaries. Panel A shows ridge-probe $R^2$ for decoding
  posterior mean parameters across tasks, with thin lines for tasks and thick
  lines for task-family means. Panel B shows marginal-quantile probe pinball
  loss normalized by empirical-reference-quantile pinball loss on the same
  reference observations; $1\times$ is the reference target.}
  \label{fig:probe_main_summary}
\end{figure}

\Cref{fig:probe_main_summary} measures whether frozen summaries contain
linearly accessible information about posterior marginals before fitting the
density head. We fit ridge probes from layer-wise summaries to posterior mean
parameters and linear quantile probes to empirical reference-posterior marginal
quantiles. The posterior-mean probe improves across encoder depth, and the
marginal-quantile probe moves toward the empirical-reference pinball target on
many tasks. The raw-observation ablation shows the same trend
(\cref{tab:quantile_raw_ablation}): PFN summaries give higher
predicted-versus-reference quantile correlations than standardized raw
observations on SLCP, SIR, Lotka--Volterra, AR(1), Bernoulli GLM variants, and
several distractor settings. These results show that the summaries retain useful
one-dimensional information about posterior location, scale, and quantiles.

The quantile diagnostics also vary by task. The pinball ratio is near the
empirical-reference target on the Gaussian and two-moons tasks. SIR,
Lotka--Volterra, OU, solar dynamo, and AR(1) have larger pinball ratios
(\cref{tab:quantile_raw_ablation}). The summaries often expose marginal
posterior signal even when final joint C2ST remains elevated.

\subsection{Joint error after marginal normalization}
\label{sec:diagnostic:joint}

\begin{figure}[t]
  \centering
  \includegraphics[width=0.9\linewidth]{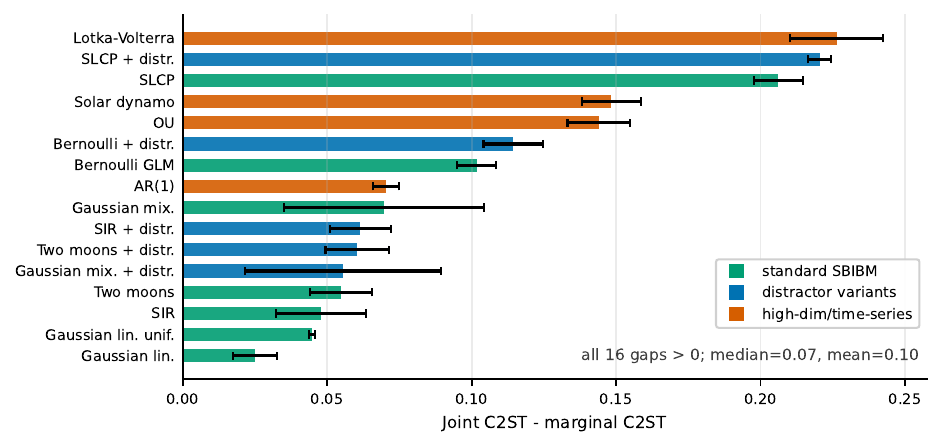}
  \caption{Marginal-vs-joint C2ST gap for PFN-NPE at
  $n_{\text{train}}=10{,}000$. Bars show joint C2ST minus marginal C2ST,
  averaged over available seeds. A zero gap corresponds to equal joint and mean
  marginal C2ST.}
  \label{fig:c2st_decomposition}
\end{figure}

\Cref{fig:c2st_decomposition} summarizes the posterior-sample diagnostic.
PFN-NPE has higher joint C2ST than marginal C2ST on every reported task. The
joint-minus-marginal gap ranges from $0.025$ on Gaussian linear to $0.226$ on
Lotka--Volterra, with large gaps on SLCP ($0.206$), SLCP with distractors
($0.220$), OU ($0.144$), and solar dynamo ($0.148$)
(\cref{tab:c2st_marginal_joint_gap}). These gaps locate recurring error beyond
average one-dimensional marginal mismatch.

Rank-space C2ST gives another view of the same pattern. The rank transform
normalizes each parameter coordinate by the pooled empirical CDF before
applying C2ST. This reduces one-dimensional marginal effects while retaining
dependence information. Across the task suite, rank C2ST remains close to joint
C2ST and well above the marginal score on high-gap tasks
(\cref{tab:c2st_marginal_joint_gap}). This pattern links accessible marginal
information with elevated joint C2ST after fitting the sample distribution.

\subsection{Parameter-level heterogeneity}
\label{sec:diagnostic:parameter_heterogeneity}

The cross-$\theta$ probe in the appendix asks how parameter information is
distributed across parameter-specific summary chunks. Let
\(h_{\mathrm{raw}}(x)=[e_1(x),\ldots,e_{D_\theta}(x)]\) denote the
per-parameter concatenated TabPFN embedding at a given layer. For each source
chunk \(i\) and target parameter \(j\), we fit a ridge probe,
\(\hat{\theta}_j = a_{ij}^{\top} e_i(x) + b_{ij}\), on training simulations
and report validation \(R^2_{ij}\). The matched score is \(R^2_{jj}\); the
off-mean score averages \(R^2_{ij}\) over \(i\neq j\).
\Cref{tab:probe_parameter_roles} reports these quantities for the final encoder
layer, averaged across seeds.

The final-layer probe table shows heterogeneous accessibility across simulator
parameters. In SLCP, the two location parameters are decodable from their
matched chunks with \(R^2 \approx 0.60\), the two scale parameters have
near-zero \(R^2\), and the correlation-like parameter is partly decodable in
the base task (\(R^2=0.38\)). In AR(1), \(\rho\) is strongly localized to its
matched chunk (\(0.82\) vs.\ \(0.19\)), and \(\log\sigma\) is accessible from
both chunks. The same frozen summary construction can therefore produce
localized, distributed, and weakly linear parameter signals.

The frozen summaries make some coordinates linearly accessible from their
matched chunks, make others accessible across chunks, and leave some weak under
this linear probe. The C2ST decomposition shows where sample error remains:
after fitting the conditional density estimator, PFN-NPE samples still show
discrepancies in dependence and parameter interactions. Overall, the
diagnostics show that PFN-NPE carries useful one-dimensional posterior
information into posterior estimation, with remaining error concentrated in
multivariate structure.

% \input{sections/closing_loop}
% \input{sections/copula}
% !TEX root = ../main.tex
\section{Discussion}
\label{sec:discussion}

PFN-NPE treats a pretrained TabPFN as a reusable summary network and pairs the
embeddings with a task-specific posterior estimator. Across the benchmark
suite, it provides a task-dependent alternative to direct NPE-PFN inference and
learned-summary NPE. PFN-NPE performs best on mixture and nuisance-coordinate
settings, including \texttt{gaussian\_mixture},
\texttt{gaussian\_mixture\_distractors}, and
\texttt{two\_moons\_distractors}. Direct NPE-PFN gives the lowest
fixed-budget joint C2ST on many standard benchmark tasks. These results support
PFN-NPE as a modular summary-based method with specific strengths, especially
when frozen TabPFN summaries provide useful input to a downstream density
estimator.

The diagnostics explain this pattern. Linear probes recover posterior means and
marginal quantiles from frozen TabPFN summaries across a range of tasks, showing
that the representations often contain useful one-dimensional posterior
information. The joint-minus-marginal C2ST gap and rank-space C2ST locate
remaining error in the multivariate sampled posterior. Together, these results
suggest that PFN-derived summaries often preserve marginal posterior signal,
while dependence error remains in the final samples. This matches the TabPFN
pretraining objective: TabPFN is trained on scalar supervised prediction tasks
drawn from a prior over structural causal models
\citep{hollmann2023tabpfn,hollmann2025tabpfn}. In the PFN view of
\citet{nagler2023statistical}, the learned predictor approximates a posterior
predictive distribution under this pretraining prior. SBI tasks whose summaries
resemble scalar prediction can align with that prior; tasks that require
accurate multivariate dependence can expose a mismatch between the summaries
and the target posterior. The diagnostics support this view for the fitted
PFN-NPE system, but they do not fully separate representation error from
density-estimator error.

\subsection{Limitations}
\label{sec:discussion:limits}

The benchmark suite covers standardized SBIBM simulators, nuisance-coordinate
variants, and three structured time-series tasks under a common evaluation
protocol. It remains small relative to the SBI problem space and emphasizes
accessible benchmarks. Broader coverage of expensive mechanistic simulators,
higher-dimensional parameter spaces, and controlled dependence structures would
clarify where PFN-NPE is most useful. The evaluation also relies heavily on
C2ST. C2ST measures sample-level discrepancy, but source attribution requires
additional diagnostics. Appendix~\ref{app:slcp_failure_diagnostics} illustrates
this point through a detailed SLCP case study. Finally, the default PFN-NPE
implementation fixes the PCA-64 representation and one normalizing-flow
configuration. Other compression choices and posterior estimators may change
the ranking.

\subsection{Future work}
\label{sec:discussion:future}

The results point to several extensions. One direction is to make joint
structure explicit in PFN-NPE, for example by coupling
parameter-specific embeddings, adding summaries that target simulator
dependence, or fine-tuning a tabular foundation model on parameter-prediction
objectives. These experiments would test whether the marginal-vs-joint gap
comes from the frozen representation, the posterior estimator, or their
interaction. A second direction is a broader benchmark suite with gradual
changes in multimodality, dependence strength, observation dimension, and
simulation cost. Appendix~\ref{app:wall_clock} reports wall-clock and anytime
Pareto analyses for two moons, SLCP, and AR(1), treating runtime for ten
posterior queries as an empirical constraint. Extending these timing sweeps to
larger-budget simulators, including problems such as the pyloric network
benchmark studied by \citet{vetter2025npepfn}, would clarify the tradeoff
between simulation budget and wall-clock budget. Finally, richer versions of PFN-NPE could prove fruitful. For example, one could explore gains from pairing TabPFN embeddings with hand-crafted simulator summary statistics, combining per-dimension summary embeddings in different ways (as opposed to PCA), or using a more expressive inference head, like a transformer-based diffusion model, that may be better suited to multimodal or strongly dependent posteriors.

% !TEX root = ../main.tex
% The `ack` environment is provided by neurips_2026.sty and is hidden in
% anonymized submissions automatically. Only fill in for the final/preprint version.
\begin{ack}
We thank the maintainers of TabPFN, SBIBM, sbi, BayesFlow, NPE-PFN, Zuko, uv, and the scientific Python ecosystem for making their software available.
\end{ack}

\newpage

\bibliographystyle{plainnat}
\bibliography{references}

\clearpage
\appendix
% \input{sections/appendix}
% !TEX root = ../main.tex
\section{Additional Method Details}
\subsection{Pseudocode: PFN-NPE}
\begin{algorithm}[th]
\caption{PFN-NPE}
\label{alg:pfn-npe}
\begin{algorithmic}[1]
\Require Observation $\mathbf{x}_o$, training simulations $\mathcal{D} = \{(\boldsymbol{\theta}_i, \mathbf{x}_i)\}_{i=1}^{n}$, frozen TabPFN encoder $h_\psi(\cdot)$, dimensionality reduction $g$ (e.g.\ PCA to 64 dimensions), inference head $q_\phi$

\vspace{0.5em}
\Statex \textbf{\textit{Extract summary embeddings}}

\State Sample in-context subset $\mathcal{C} \subseteq \mathcal{D}$ with $|\mathcal{C}| = \min(1000, n)$ uniformly without replacement

\For{$d = 1, \ldots, D_\theta$}
    \State $\mathcal{C}_d = \{(\mathbf{x}_i, \theta_{i,d}) : (\boldsymbol{\theta}_i, \mathbf{x}_i) \in \mathcal{C}\}$
    \For{each $(\boldsymbol{\theta}_i, \mathbf{x}_i) \in \mathcal{D}$}
        \State $h_d(\mathbf{x}_i) \leftarrow$ final-layer embedding from $h_\psi$ with context $\mathcal{C}_d$ and query $\mathbf{x}_i$
    \EndFor
\EndFor
\For{each $(\boldsymbol{\theta}_i, \mathbf{x}_i) \in \mathcal{D}$}
    \State $h_\text{raw}(\mathbf{x}_i) \leftarrow [h_1(\mathbf{x}_i), \ldots, h_{D_\theta}(\mathbf{x}_i)]$
\EndFor

\vspace{0.5em}
\Statex \textbf{\textit{Fit transformation and inference head}}

\State Fit $g$ on $\{h_\text{raw}(\mathbf{x}_i)\}_{i=1}^n$ \Comment{$g$ is fixed after this step; TabPFN is not queried again}

\State $s(\mathbf{x}_i) \leftarrow g\bigl(h_\text{raw}(\mathbf{x}_i)\bigr)$ for each $i$ \Comment{compute fixed summary vectors}

\State Train $q_\phi$ by minimizing $\displaystyle -\frac{1}{n}\sum_{i=1}^{n} \log q_\phi\!\left(\frac{\boldsymbol{\theta}_i - \boldsymbol{\mu}_\theta}{\boldsymbol{\sigma}_\theta}\ \Big|\ s(\mathbf{x}_i)\right)$
\Comment{flow trained on fixed embeddings}

\vspace{0.5em}
\Statex \textbf{\textit{Inference}}

\For{$d = 1, \ldots, D_\theta$}
    \State $h_d(\mathbf{x}_o) \leftarrow$ final-layer embedding from $h_\psi$ with context $\mathcal{C}_d$ and query $\mathbf{x}_o$
\EndFor

\State $h_\text{raw}(\mathbf{x}_o) \leftarrow \bigl[h_1(\mathbf{x}_o),\ \ldots,\ h_{D_\theta}(\mathbf{x}_o)\bigr]$

\State $s(\mathbf{x}_o) \leftarrow g\bigl(h_\text{raw}(\mathbf{x}_o)\bigr)$

\State Draw standardized samples $\tilde{\boldsymbol{\theta}} \sim q_\phi\bigl(\cdot \mid s(\mathbf{x}_o)\bigr)$

\State Rescale samples $\boldsymbol{\theta} \leftarrow \boldsymbol{\mu}_\theta + \boldsymbol{\sigma}_\theta \odot \tilde{\boldsymbol{\theta}}$

\State \Return $\boldsymbol{\theta} = [\theta_1,\theta_2,...,\theta_{D_{\theta}}]^{\top}$ \Comment{posterior sample}

\end{algorithmic}
\end{algorithm}

\subsection{Flow training details}
\label{app:training-details}
For tasks with \(D_\theta \leq 5\), the flow uses five spline transforms,
eight spline bins, and conditioner hidden widths \([128,128]\). For
higher-dimensional parameter spaces, it uses eight transforms and hidden widths
\([256,256]\). We train this head by minimizing negative log likelihood,
\[
  \mathcal{L}(\phi)
  =
  -\frac{1}{n}\sum_{i=1}^n
  \log q_\phi\!\left(
    \frac{\theta_i-\mu_\theta}{\sigma_\theta}
    \,\middle|\, s(x_i)
  \right),
\]
where \(\mu_\theta\) and \(\sigma_\theta\) are computed from the training
parameters. Optimization uses Adam with learning rate \(5\times10^{-4}\), batch
size 256, cosine learning-rate decay, gradient clipping at norm 5, early
stopping after 20 validation epochs without improvement, and at most 200
epochs.

\subsection{Task List}

\begin{table}[t]
  \centering
  \caption{SBI Task Families}
  \label{tab:task_families}
  \begin{tabular}{p{0.24\linewidth}p{0.42\linewidth}p{0.25\linewidth}}
    \toprule
    Family & Tasks & Purpose \\
    \midrule
    SBIBM reference suite &
    \texttt{two\_moons}, \texttt{gaussian\_mixture}, \texttt{slcp},
    \texttt{slcp\_distractors}, \texttt{gaussian\_linear},
    \texttt{gaussian\_linear\_uniform}, \texttt{bernoulli\_glm},
    \texttt{bernoulli\_glm\_raw}, \texttt{sir}, \texttt{lotka\_volterra} &
    Standardized comparison using established SBI tasks and reference
    posteriors. \\
    Controlled distractor variants &
    \texttt{two\_moons\_distractors},
    \texttt{gaussian\_mixture\_distractors},
    \texttt{bernoulli\_glm\_distractors}, \texttt{sir\_distractors} &
    Nuisance-coordinate stress test with a preserved target posterior. \\
    High-dimensional time series &
    \texttt{ar1\_ts\_t50}, \texttt{ou}, \texttt{solar\_dynamo} &
    Structured sequential observations with grid-based reference posterior
    samples. \\
    \bottomrule
  \end{tabular}
\end{table}

Table \ref{tab:task_families} lists the task families used in the empirical
comparison. The SBIBM tasks provide the externally comparable benchmark
setting. The distractor and time-series tasks stress high-dimensional
observations. The time-series tasks and \texttt{slcp\_distractors} are not in
the SBIBM package; they are custom implementations included in the project repository referenced in \cref{app:software}.
% supplementary zip file
\subsection{Custom time-series simulators}
\label{app:custom-time-series}

The three custom time-series tasks are synthetic stress tests, not calibrated
scientific models. They test inference from structured sequential observations
with low-dimensional parameters and grid-based reference posteriors. The
Ornstein--Uhlenbeck and solar-dynamo tasks are inspired by the high-dimensional
SBI settings in \citet{dirmeier2023ssnl}, but we use the custom definitions,
priors, reference observations, and posterior grids described below.

\paragraph{AR(1).}
The \texttt{ar1\_ts\_t50} task has parameter
\(\theta=(\rho,\log\sigma)\) and observation
\(x=(x_1,\ldots,x_{50})\). The prior is
\[
  \rho \sim \operatorname{Uniform}(-0.95,0.95),
  \qquad
  \log\sigma \sim \operatorname{Uniform}(\log 0.05,\log 2).
\]
Given \(\theta\), the simulator draws
\[
  x_1 \sim \mathcal{N}\!\left(0, \frac{\sigma^2}{1-\rho^2}\right),
  \qquad
  x_t = \rho x_{t-1} + \epsilon_t,\quad
  \epsilon_t \sim \mathcal{N}(0,\sigma^2),
\]
for \(t=2,\ldots,50\). The implementation clamps \(1-\rho^2\) below by
\(10^{-4}\) when computing the stationary initial variance.

\paragraph{Ornstein--Uhlenbeck.}
The \texttt{ou} task has parameter \(\theta=(\alpha,\beta,\sigma)\),
observation length \(100\), and time step \(\Delta t=0.1\). The prior is
uniform on \([0,10]\times[0,5]\times[0,2]\). The simulator uses the exact
Gaussian transition
\[
  x_{t+1}\mid x_t
  \sim
  \mathcal{N}\!\left(
    \alpha + (x_t-\alpha)\exp(-\beta\Delta t),
    \frac{\sigma^2\{1-\exp(-2\beta\Delta t)\}}{2\beta}
  \right),
\]
with initial state
\[
  x_1 \sim \mathcal{N}\!\left(\alpha,\frac{\sigma^2}{2\beta}\right).
\]
For numerical stability, the implementation evaluates these expressions with
\(\beta\) bounded below by \(10^{-6}\).

\paragraph{Solar-dynamo-inspired toy model.}
The \texttt{solar\_dynamo} task has parameter
\(\theta=(\alpha_{\min},\alpha_{\mathrm{range}},\epsilon_{\max})\) and
observation length \(100\). The prior is uniform on
\([0.9,1.4]\times[0.05,0.25]\times[0.02,0.15]\). Starting from fixed
initial state \(p_0=1\), the simulator iterates
\[
  p_{t+1} = \alpha_t f(p_t)p_t + \epsilon_t,
  \qquad
  \alpha_t \sim
  \operatorname{Uniform}(\alpha_{\min},
  \alpha_{\min}+\alpha_{\mathrm{range}}),
  \qquad
  \epsilon_t \sim \operatorname{Uniform}(0,\epsilon_{\max}),
\]
where
\[
  f(p)
  =
  \frac{1}{2}\left\{1+\operatorname{erf}\!\left(\frac{p-0.6}{0.2}\right)\right\}
  \left\{1-\operatorname{erf}\!\left(\frac{p-1.0}{0.8}\right)\right\}.
\]
This task is only intended as a nonlinear sequential benchmark; we do not use
it as a calibrated model of solar-cycle dynamics.

\paragraph{Reference posteriors.}
For each custom time-series task, we generate ten fixed reference observations
from prior draws using seed \(12345\) and cache \(10{,}000\) reference posterior
samples per observation. The AR(1) task evaluates the exact likelihood on a
\(401\times301\) grid over \((\rho,\log\sigma)\). The OU and solar-dynamo
tasks evaluate their analytic transition likelihoods on \(80^3\) parameter
grids. Posterior samples are drawn by normalizing the grid log posterior and
sampling grid cells with replacement.

\subsection{Computational Resources}
\label{app:computational-resources}

The benchmark, diagnostic, ablation, case-study, and amortized wall-clock
experiments ran on an Ubuntu workstation with one NVIDIA RTX 3090 Ti GPU with
24 GB of VRAM.

The anytime Pareto frontier wall-clock analysis in
\cref{fig:anytime_pareto_joint_c2st} ran on a local Linux \texttt{x86\_64}
workstation with 16 physical CPU cores, 22 logical CPU cores, 61.0 GB of RAM,
and one NVIDIA RTX 3500 Ada Generation Laptop GPU with 12,282 MB of VRAM
\mbox{(CUDA compute capability 8.9, driver 595.58.03)}.

\subsection{Software}
\label{app:software}

%in the  and supplementary zip file
The code for this project is available in the \texttt{epickens/pfn-npe} GitHub repository. Package dependencies can be
installed from the pyproject requirements using \texttt{uv}.

\clearpage
\section{Additional results}
\label{app:additional}

\subsection{Wall-clock comparison}
\label{app:wall_clock}

The timing appendix reports implementation-level runtimes for the local timing
suite. Train time includes simulation plus task-specific fitting or embedding
steps. Sampling time measures \(1000\) posterior draws for each of ten reference
observations. These measurements depend on hardware and implementation, so they
describe this implementation rather than the methods in general.

\begin{table}[ht]
  \centering
  \caption{Wall-clock cost on the timing suite.}
  \label{tab:wall_clock}
  % Auto-generated by scripts/wall_clock_table.py. Do not edit by hand.
\begin{tabular}{llrrr}
  \toprule
  Task & Method & Train (s) & Sample (s) & Total (s) \\
  \midrule
  \texttt{slcp} & NPE-PFN & $0 \pm 0$ & $253 \pm 0$ & $256 \pm 0$ \\
   & PFN-NPE (NSF) & $43 \pm 4$ & $0 \pm 0$ & $46 \pm 4$ \\
   & Learned summary + NSF & $41 \pm 7$ & $0 \pm 0$ & $41 \pm 7$ \\
   & Vanilla NPE & $70 \pm 5$ & $0 \pm 0$ & $73 \pm 5$ \\
   & PyMC HMC & $0 \pm 0$ & $43 \pm 1$ & $46 \pm 1$ \\
  \midrule
  \texttt{two\_moons\_distractors} & NPE-PFN & $0 \pm 0$ & $396 \pm 26$ & $400 \pm 26$ \\
   & PFN-NPE (NSF) & $73 \pm 4$ & $0 \pm 0$ & $75 \pm 4$ \\
   & Learned summary + NSF & $19 \pm 2$ & $0 \pm 0$ & $19 \pm 2$ \\
   & Vanilla NPE & $21 \pm 1$ & $0 \pm 0$ & $24 \pm 1$ \\
   & PyMC HMC & — & — & — \\
  \midrule
  \texttt{gaussian\_linear} & NPE-PFN & $0 \pm 0$ & $331 \pm 0$ & $334 \pm 0$ \\
   & PFN-NPE (NSF) & $27 \pm 1$ & $0 \pm 0$ & $30 \pm 1$ \\
   & Learned summary + NSF & $20 \pm 1$ & $0 \pm 0$ & $20 \pm 1$ \\
   & Vanilla NPE & $17 \pm 1$ & $0 \pm 0$ & $20 \pm 1$ \\
   & PyMC HMC & $0 \pm 0$ & $4 \pm 0$ & $7 \pm 0$ \\
  \bottomrule
\end{tabular}

\end{table}

\Cref{tab:wall_clock} separates setup cost from repeated posterior-query cost
at \(n_{\text{train}}=10{,}000\). NPE-PFN has little task-specific fitting
cost beyond simulation and uses in-context inference for each posterior query.
PFN-NPE adds task-specific embedding and flow fitting, followed by low sampling
cost for repeated observations. PyMC HMC is included only for tasks with
tractable analytic likelihoods.

\begin{figure}[ht]
  \centering
  \includegraphics[width=\linewidth]{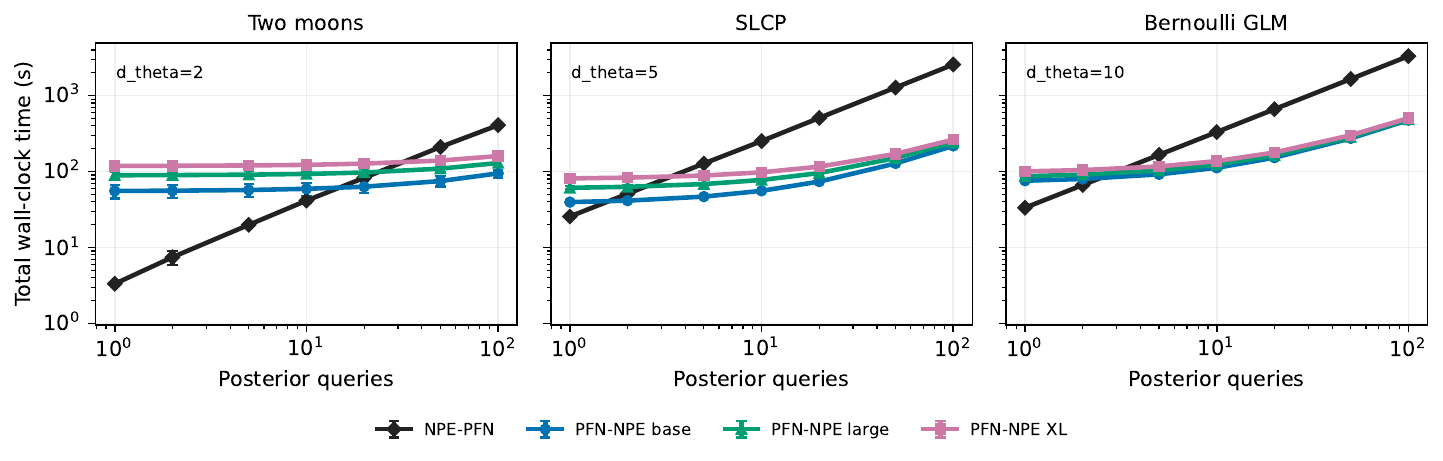}
  \caption{Amortized runtime for repeated posterior queries.}
  \label{fig:wall_clock_amortization}
\end{figure}

\Cref{fig:wall_clock_amortization} shows how total runtime changes as posterior
queries increase. Lines show mean total runtime across timing-only seeds, and
error bars show across-seed standard deviation. PFN-NPE variants pay a one-time
flow-training cost and keep low sampling cost in repeated-use settings. NPE-PFN
avoids task-specific fitting, which helps for one-off queries, but it performs
TabPFN in-context inference inside each posterior query. In the timing suite,
ten reference observations require \(253\)--\(396\) seconds of NPE-PFN sampling
time, while PFN-NPE pays \(27\)--\(73\) seconds up front and its repeated
sampling cost rounds to zero at table precision (\cref{tab:wall_clock}). For
many observed datasets from the same simulator setup, this per-query cost can
dominate total runtime even when NPE-PFN has no training phase.

\begin{figure}[t]
  \centering
  \includegraphics[width=\linewidth]{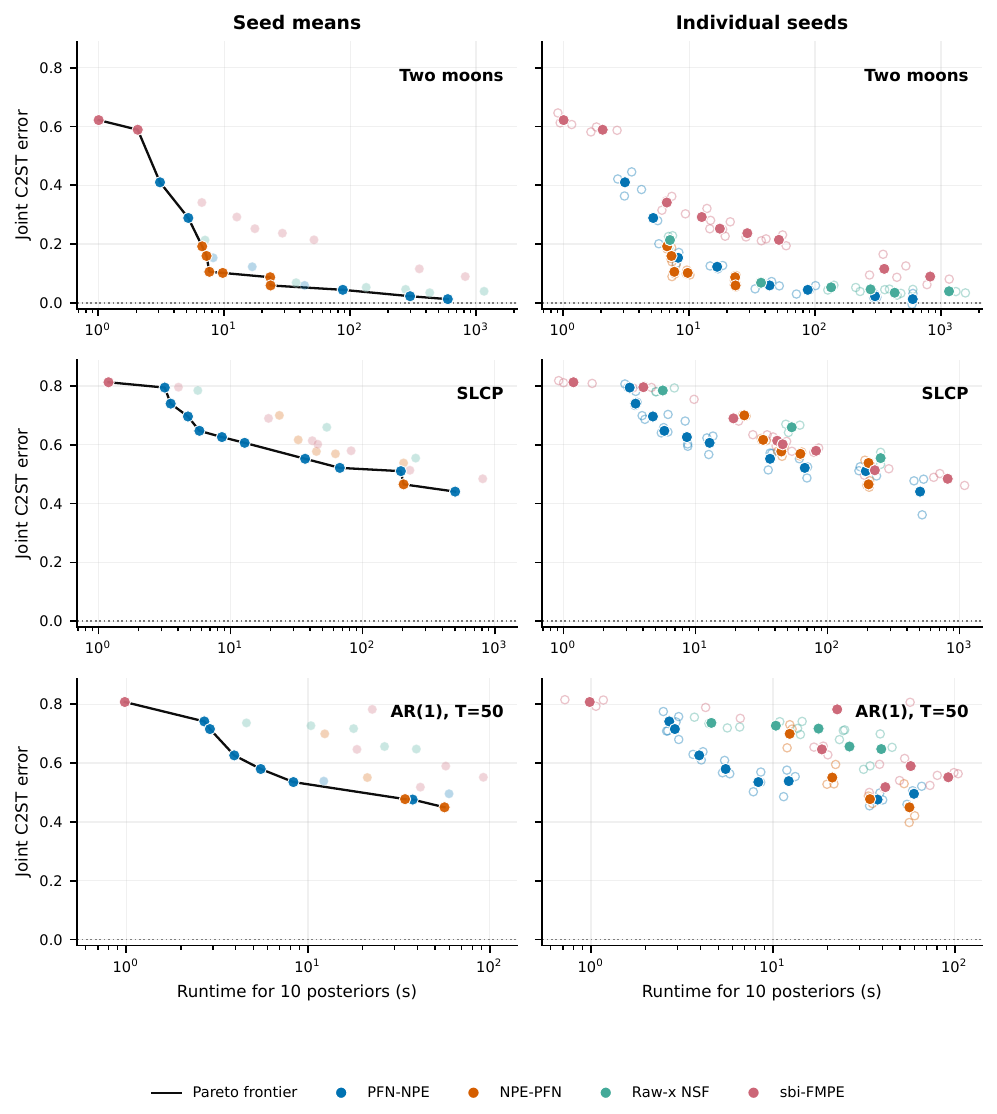}
  \caption{Anytime quality-runtime comparison on the timing suite.}
  \label{fig:anytime_pareto_joint_c2st}
\end{figure}

\Cref{fig:anytime_pareto_joint_c2st} evaluates posterior quality under a
wall-clock constraint on two moons, SLCP, and AR(1). Rows correspond to tasks.
The left column shows seed means for each method and training budget, with
non-dominated points connected by the Pareto frontier. The right column
overlays individual seeds and seed means. Runtime includes posterior sampling
for ten reference observations. The vertical axis reports joint C2ST error,
\(2(\mathrm{C2ST}_{\mathrm{joint}} - 0.5)\), so lower-left points have lower
runtime and lower posterior discrepancy.

The frontier depends on the task. On two moons, NPE-PFN reaches low joint C2ST
error within tens of seconds, while higher-budget PFN-NPE reaches the lowest
observed aggregate error at longer runtimes. On SLCP, PFN-NPE contributes most
of the intermediate frontier, with NPE-PFN and PFN-NPE defining the
high-runtime end. On AR(1), all methods retain high joint C2ST error; PFN-NPE
is competitive at short runtimes, and NPE-PFN gives the lowest observed
aggregate error among the plotted configurations. The seed-level panels show
moderate seed variation with the same task-level pattern.
 
\clearpage
\subsection{Full per-task joint C2ST table}
\label{app:c2st_full}

The full \(10{,}000\)-simulation table expands the compact family summary in
the main text to individual tasks. Entries are joint C2ST means with
across-seed standard deviations; lower values indicate posterior samples closer
to the reference samples, and bold marks the lowest mean in each row. The
external-baseline table uses the standard SBIBM benchmark rows and reports
rounded joint C2ST values at the same simulation budget. PFN-NPE is recomputed
from local posterior samples; the other shipped baselines use the SBIBM
benchmark results. FMPE appears where a local result is available.

\begin{table}[t]
  \centering
  \caption{Per-task joint C2ST at \(n_{\text{train}}=10{,}000\).}
  \label{tab:c2st_joint_full}
  % Auto-generated by scripts/aggregate_c2st_table.py. Do not edit by hand.
\begin{tabular}{lccc}
  \toprule
  Task & PFN-NPE & Learned summary + NSF & NPE-PFN \\
  \midrule
  ar1\_ts\_t50 & $0.783\,\pm\,0.02$ & $\mathbf{0.692}\,\pm\,0.01$ & $0.728\,\pm\,0.01$ \\
  bernoulli\_glm & $0.657\,\pm\,0.00$ & $0.701\,\pm\,0.02$ & $\mathbf{0.559}\,\pm\,0.01$ \\
  bernoulli\_glm\_distractors & $0.680\,\pm\,0.01$ & $0.763\,\pm\,0.01$ & $\mathbf{0.562}\,\pm\,0.01$ \\
  bernoulli\_glm\_raw & $0.707\,\pm\,0.02$ & --- & $\mathbf{0.593}\,\pm\,0.01$ \\
  gaussian\_linear & $0.553\,\pm\,0.01$ & $0.545\,\pm\,0.00$ & $\mathbf{0.510}\,\pm\,0.00$ \\
  gaussian\_linear\_uniform & $0.588\,\pm\,0.00$ & $0.551\,\pm\,0.01$ & $\mathbf{0.509}\,\pm\,0.01$ \\
  gaussian\_mixture & $\mathbf{0.625}\,\pm\,0.05$ & $0.626\,\pm\,0.02$ & $0.704\,\pm\,0.01$ \\
  gaussian\_mixture\_distractors & $\mathbf{0.605}\,\pm\,0.04$ & $0.762\,\pm\,0.01$ & $0.739\,\pm\,0.00$ \\
  lotka\_volterra & $0.931\,\pm\,0.01$ & $0.944\,\pm\,0.01$ & $\mathbf{0.832}\,\pm\,0.01$ \\
  ou & $\mathbf{0.866}\,\pm\,0.01$ & $0.910\,\pm\,0.00$ & $0.890\,\pm\,0.00$ \\
  sir & $0.608\,\pm\,0.02$ & $0.647\,\pm\,0.02$ & $\mathbf{0.544}\,\pm\,0.01$ \\
  sir\_distractors & $0.626\,\pm\,0.01$ & $0.645\,\pm\,0.01$ & $\mathbf{0.563}\,\pm\,0.01$ \\
  slcp & $0.826\,\pm\,0.01$ & $0.811\,\pm\,0.01$ & $\mathbf{0.810}\,\pm\,0.00$ \\
  slcp\_distractors & $0.861\,\pm\,0.01$ & --- & $\mathbf{0.836}\,\pm\,0.01$ \\
  solar\_dynamo & $0.868\,\pm\,0.01$ & $0.944\,\pm\,0.00$ & $\mathbf{0.857}\,\pm\,0.00$ \\
  two\_moons & $0.581\,\pm\,0.02$ & $\mathbf{0.561}\,\pm\,0.01$ & $0.602\,\pm\,0.00$ \\
  two\_moons\_distractors & $\mathbf{0.582}\,\pm\,0.02$ & $0.917\,\pm\,0.00$ & $0.603\,\pm\,0.00$ \\
  \bottomrule
\end{tabular}

\end{table}

\begin{table}[t]
  \centering
  \caption{SBIBM external-baseline comparison at
  \(n_{\text{train}}=10{,}000\).}
  \label{tab:external_baselines_10k}
  \resizebox{\linewidth}{!}{% Auto-generated by scripts/paper_comparison_tables.py. Do not edit by hand.
\begin{tabular}{lccccccc}
  \toprule
  Task & PFN-NPE & NPE & NLE & NRE & FMPE & SNPE & SNLE \\
  \midrule
  Two moons & $0.58$ & $0.61$ & $0.71$ & $0.76$ & --- & $\mathbf{0.55}$ & $0.57$ \\
  Gaussian mixture & $0.62$ & $0.66$ & $0.73$ & $0.75$ & --- & $\mathbf{0.58}$ & $0.70$ \\
  Gaussian linear & $0.55$ & $0.55$ & $0.55$ & $0.56$ & --- & $\mathbf{0.54}$ & $0.55$ \\
  Bernoulli GLM & $0.66$ & $0.68$ & $0.60$ & $0.81$ & --- & $0.61$ & $\mathbf{0.55}$ \\
  SLCP & $0.82$ & $0.90$ & $0.77$ & $0.95$ & $0.83$ & $0.84$ & $\mathbf{0.71}$ \\
  SIR & $0.61$ & $0.68$ & $0.75$ & $0.77$ & --- & $\mathbf{0.56}$ & $0.74$ \\
  Lotka-Volterra & $0.93$ & $1.00$ & $0.96$ & $1.00$ & --- & $0.95$ & $\mathbf{0.74}$ \\
  \bottomrule
\end{tabular}
}
\end{table}

\subsection{High-budget performance tables}
\label{app:high_budget_performance}

The high-budget tables repeat the main comparison at
\(n_{\text{train}}=100{,}000\), where available. The compact table reports
which local method has the lowest mean joint C2ST within each task family. The
per-task table reports local-method means and across-seed standard deviations.
The external-baseline table gives the corresponding standard SBIBM comparison
with rounded joint C2ST values. Dashes mark configurations without a completed
local run at this budget.

\begin{table}[t]
  \centering
  \caption{Compact \(100{,}000\)-simulation summary by task family.}
  \label{tab:compact_family_summary_100k}
  \resizebox{\linewidth}{!}{% Auto-generated by scripts/paper_comparison_tables.py. Do not edit by hand.
\begin{tabular}{lrrrrcc}
  \toprule
  Task family & \# tasks & PFN-NPE best & Learned summary best & NPE-PFN best & Median PFN-NPE & Median best \\
  \midrule
  SBIBM reference & 10 & 2 & 1 & 7 & $0.564$ & $0.536$ \\
  Distractor variants & 4 & 2 & 0 & 2 & $0.559$ & $0.551$ \\
  Time series & 3 & 1 & 1 & 1 & $0.864$ & $0.859$ \\
  \bottomrule
\end{tabular}
}
\end{table}

\begin{table}[t]
  \centering
  \caption{Per-task joint C2ST at \(n_{\text{train}}=100{,}000\).}
  \label{tab:c2st_joint_100k}
  % Auto-generated by scripts/aggregate_c2st_table.py. Do not edit by hand.
\begin{tabular}{lccc}
  \toprule
  Task & PFN-NPE & Learned summary + NSF & NPE-PFN \\
  \midrule
  ar1\_ts\_t50 & $0.783\,\pm\,0.01$ & $\mathbf{0.581}\,\pm\,0.02$ & $0.724\,\pm\,0.01$ \\
  bernoulli\_glm & $0.591\,\pm\,0.01$ & $0.586\,\pm\,0.01$ & $\mathbf{0.532}\,\pm\,0.00$ \\
  bernoulli\_glm\_distractors & $0.594\,\pm\,0.01$ & $0.633\,\pm\,0.01$ & $\mathbf{0.554}\,\pm\,0.00$ \\
  bernoulli\_glm\_raw & --- & --- & $\mathbf{0.567}\,\pm\,0.01$ \\
  gaussian\_linear & $0.532\,\pm\,0.00$ & $0.522\,\pm\,0.00$ & $\mathbf{0.509}\,\pm\,0.00$ \\
  gaussian\_linear\_uniform & --- & $0.521\,\pm\,0.00$ & $\mathbf{0.515}\,\pm\,0.00$ \\
  gaussian\_mixture & $\mathbf{0.541}\,\pm\,0.01$ & $0.551\,\pm\,0.01$ & $0.568\,\pm\,0.00$ \\
  gaussian\_mixture\_distractors & $\mathbf{0.552}\,\pm\,0.01$ & $0.629\,\pm\,0.01$ & $0.741\,\pm\,0.00$ \\
  lotka\_volterra & $0.861\,\pm\,0.00$ & $0.828\,\pm\,0.00$ & $\mathbf{0.690}\,\pm\,0.01$ \\
  ou & $\mathbf{0.864}\,\pm\,0.01$ & $0.900\,\pm\,0.01$ & $0.907\,\pm\,0.00$ \\
  sir & $0.564\,\pm\,0.02$ & $0.573\,\pm\,0.03$ & $\mathbf{0.518}\,\pm\,0.00$ \\
  sir\_distractors & $0.565\,\pm\,0.01$ & $0.599\,\pm\,0.01$ & $\mathbf{0.549}\,\pm\,0.00$ \\
  slcp & $0.732\,\pm\,0.02$ & $\mathbf{0.680}\,\pm\,0.01$ & $0.708\,\pm\,0.01$ \\
  slcp\_distractors & --- & --- & $\mathbf{0.777}\,\pm\,0.00$ \\
  solar\_dynamo & $0.864\,\pm\,0.01$ & $0.922\,\pm\,0.01$ & $\mathbf{0.859}\,\pm\,0.01$ \\
  two\_moons & $\mathbf{0.527}\,\pm\,0.01$ & $0.529\,\pm\,0.00$ & $0.557\,\pm\,0.00$ \\
  two\_moons\_distractors & $\mathbf{0.527}\,\pm\,0.01$ & $0.602\,\pm\,0.01$ & $0.604\,\pm\,0.00$ \\
  \bottomrule
\end{tabular}

\end{table}

\begin{table}[t]
  \centering
  \caption{SBIBM external-baseline comparison at
  \(n_{\text{train}}=100{,}000\).}
  \label{tab:external_baselines_100k}
  \resizebox{\linewidth}{!}{% Auto-generated by scripts/paper_comparison_tables.py. Do not edit by hand.
\begin{tabular}{lcccccc}
  \toprule
  Task & PFN-NPE & NPE & NLE & NRE & SNPE & SNLE \\
  \midrule
  Two moons & $\mathbf{0.53}$ & $0.54$ & $0.67$ & $0.63$ & $0.53$ & $0.58$ \\
  Gaussian mixture & $0.54$ & $0.56$ & $0.76$ & $0.73$ & $\mathbf{0.53}$ & $0.62$ \\
  Gaussian linear & $0.53$ & $\mathbf{0.51}$ & $0.51$ & $0.54$ & $0.51$ & $0.52$ \\
  Bernoulli GLM & $0.59$ & $0.56$ & $0.54$ & $0.75$ & $0.52$ & $\mathbf{0.52}$ \\
  SLCP & $0.73$ & $0.83$ & $0.70$ & $0.92$ & $0.67$ & $\mathbf{0.58}$ \\
  SIR & $\mathbf{0.56}$ & $0.58$ & $0.73$ & $0.69$ & $0.57$ & $0.65$ \\
  Lotka-Volterra & $0.86$ & $0.98$ & $0.95$ & $1.00$ & $0.93$ & $\mathbf{0.70}$ \\
  \bottomrule
\end{tabular}
}
\end{table}

\subsection{Marginal-vs-joint C2ST decomposition}
\label{app:c2st_marginal_joint_gap}

The C2ST decomposition reports the PFN-NPE posterior-sample diagnostics used
in \cref{sec:diagnostic}. Joint C2ST applies the classifier test to the full
parameter vector. Marginal C2ST averages one-dimensional tests over parameter
coordinates. Rank C2ST applies a marginal rank transform before testing the
joint sample, retaining sensitivity to dependence structure. The final column
reports joint C2ST minus marginal C2ST. Values near \(0.5\) indicate smaller
sample discrepancies.

\begin{table}[t]
  \centering
  \caption{PFN-NPE C2ST decomposition.}
  \label{tab:c2st_marginal_joint_gap}
  \resizebox{\linewidth}{!}{% Auto-generated by scripts/plot_c2st_marginal_joint_gap.py. Do not edit by hand.
\begin{tabular}{lrrrrr}
  \toprule
  Task & $n$ & Joint & Marginal & Rank & Joint - marg. \\
  \midrule
  Gaussian linear & 4 & 0.553 $\pm$ 0.01 & 0.528 $\pm$ 0.00 & 0.550 $\pm$ 0.01 & 0.025 $\pm$ 0.01 \\
  Gaussian linear uniform & 3 & 0.588 $\pm$ 0.00 & 0.543 $\pm$ 0.00 & 0.584 $\pm$ 0.00 & 0.045 $\pm$ 0.00 \\
  SIR & 5 & 0.608 $\pm$ 0.02 & 0.560 $\pm$ 0.01 & 0.606 $\pm$ 0.02 & 0.048 $\pm$ 0.02 \\
  Two moons & 5 & 0.581 $\pm$ 0.02 & 0.526 $\pm$ 0.01 & 0.614 $\pm$ 0.02 & 0.055 $\pm$ 0.01 \\
  Gaussian mixture + distractors & 5 & 0.605 $\pm$ 0.04 & 0.550 $\pm$ 0.01 & 0.601 $\pm$ 0.04 & 0.055 $\pm$ 0.03 \\
  Two moons + distractors & 5 & 0.582 $\pm$ 0.02 & 0.522 $\pm$ 0.01 & 0.614 $\pm$ 0.02 & 0.060 $\pm$ 0.01 \\
  SIR + distractors & 5 & 0.626 $\pm$ 0.01 & 0.564 $\pm$ 0.01 & 0.622 $\pm$ 0.01 & 0.061 $\pm$ 0.01 \\
  Gaussian mixture & 5 & 0.625 $\pm$ 0.05 & 0.555 $\pm$ 0.02 & 0.621 $\pm$ 0.05 & 0.070 $\pm$ 0.03 \\
  AR(1) time series & 3 & 0.783 $\pm$ 0.02 & 0.713 $\pm$ 0.01 & 0.780 $\pm$ 0.01 & 0.070 $\pm$ 0.00 \\
  Bernoulli GLM & 5 & 0.657 $\pm$ 0.00 & 0.556 $\pm$ 0.01 & 0.651 $\pm$ 0.01 & 0.102 $\pm$ 0.01 \\
  Bernoulli GLM + distractors & 5 & 0.680 $\pm$ 0.01 & 0.566 $\pm$ 0.00 & 0.676 $\pm$ 0.01 & 0.114 $\pm$ 0.01 \\
  Ornstein-Uhlenbeck & 3 & 0.866 $\pm$ 0.01 & 0.722 $\pm$ 0.00 & 0.880 $\pm$ 0.01 & 0.144 $\pm$ 0.01 \\
  Solar dynamo & 3 & 0.868 $\pm$ 0.01 & 0.720 $\pm$ 0.01 & 0.870 $\pm$ 0.01 & 0.148 $\pm$ 0.01 \\
  SLCP & 5 & 0.826 $\pm$ 0.01 & 0.620 $\pm$ 0.01 & 0.833 $\pm$ 0.01 & 0.206 $\pm$ 0.01 \\
  SLCP + distractors & 5 & 0.861 $\pm$ 0.01 & 0.640 $\pm$ 0.01 & 0.863 $\pm$ 0.01 & 0.220 $\pm$ 0.00 \\
  Lotka-Volterra & 3 & 0.931 $\pm$ 0.01 & 0.705 $\pm$ 0.02 & 0.924 $\pm$ 0.01 & 0.226 $\pm$ 0.02 \\
  \bottomrule
\end{tabular}
}
\end{table}

\subsection{\texorpdfstring{Per-dimension specialization probe overview (cross-$\theta$ probe)}{Per-dimension specialization probe overview (cross-theta probe)}}
\label{sec:diagnostic:cross_theta}

The cross-\(\theta\) probe asks whether the per-parameter concatenated
embedding organizes information by target parameter coordinate. We split the
embedding into parameter-indexed chunks, then use ridge probes to decode the
matching parameter dimension or nonmatching dimensions. Here, ``off-diag''
means any parameter not placed in the target slot of the TabPFN summary
network. Across tasks, the target-slot parameter is more linearly decodable,
especially from layer 4 onward. Other ``off-diag'' parameters are often, but
not always, partly decodable from the same embeddings.

\begin{figure}[t]
  \centering
  \includegraphics[width=\linewidth]{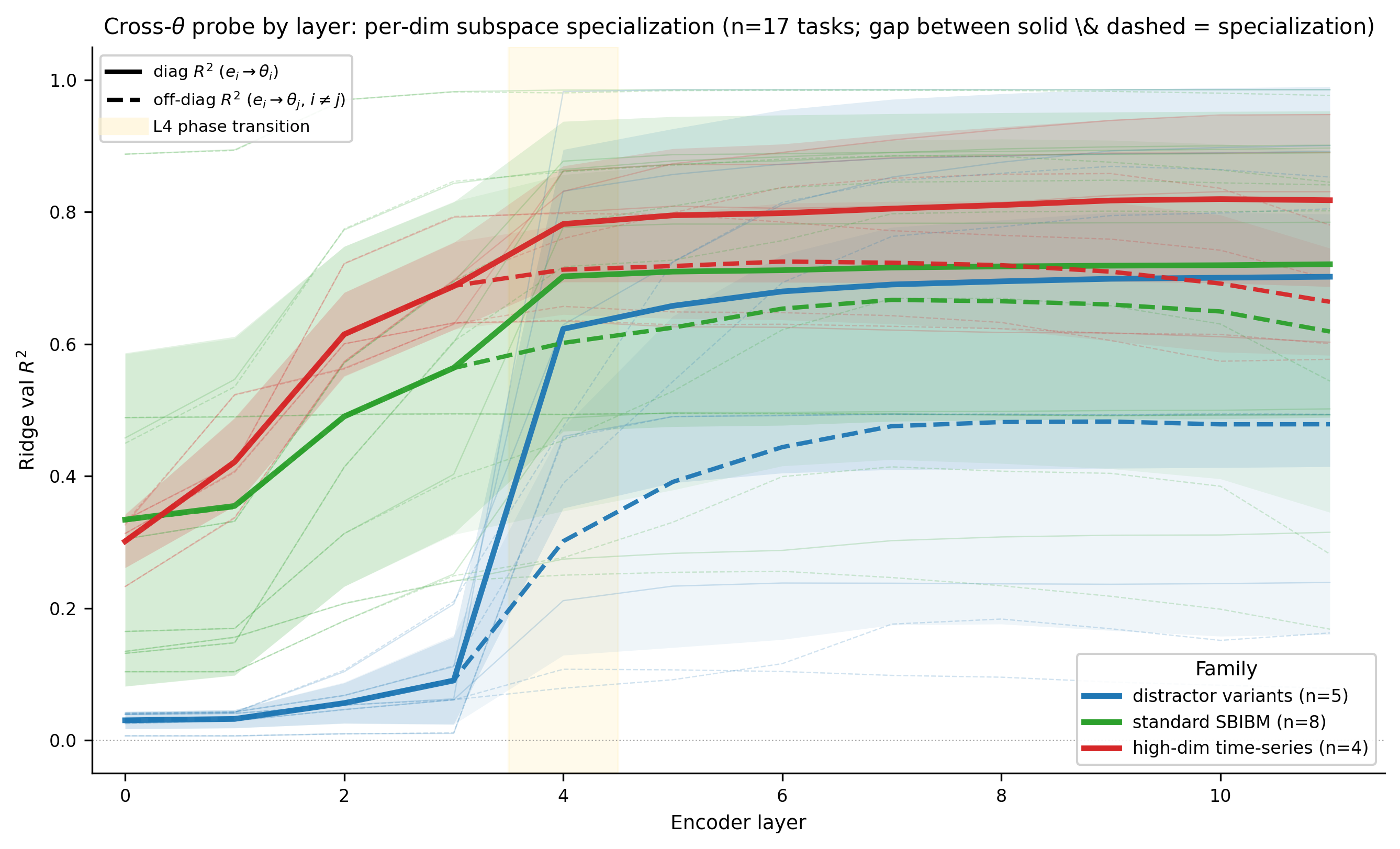}
  \caption{Layer-wise cross-\(\theta\) probe.}
  \label{fig:cross_theta_layer_trajectory}
\end{figure}

\subsection{High-budget filter diagnostic}
\label{app:p17_filter}

The filter diagnostic tests whether the standardized-Euclidean top-\(10{,}000\)
filter used by the autoregressive PFN method improves a standard NSF posterior
estimator. Entries are joint C2ST at \(n_{\text{train}}=100{,}000\), with
means and across-seed standard deviations. The \(\Delta\) column is paired by
seed and equals PFN-NPE+filter minus PFN-NPE, so positive values mean higher
C2ST after adding the filter.

\begin{table}[t]
  \centering
  \caption{High-budget top-\(10{,}000\) filter diagnostic.}
  \label{tab:p17_filter_ablation}
  \resizebox{\linewidth}{!}{% Auto-generated by scripts/p17_filter_ablation_table.py. Do not edit by hand.
\begin{tabular}{lccccc}
  \toprule
  Task & PFN-NPE & PFN-NPE+filter & $\Delta$ & AR & NPE-PFN \\
  \midrule
  \texttt{ar1\_ts\_t50} & $0.783\,\pm\,0.013$ & $0.795\,\pm\,0.027$ & $+0.012\,\pm\,0.037$ & $0.756\,\pm\,0.006$ & $\mathbf{0.724}\,\pm\,0.011$ \\
  \texttt{bernoulli\_glm} & $0.591\,\pm\,0.005$ & $0.697\,\pm\,0.005$ & $+0.106\,\pm\,0.005$ & $0.546\,\pm\,0.004$ & $\mathbf{0.532}\,\pm\,0.002$ \\
  \texttt{slcp} & $0.732\,\pm\,0.019$ & $0.793\,\pm\,0.011$ & $+0.061\,\pm\,0.029$ & $0.764\,\pm\,0.008$ & $\mathbf{0.708}\,\pm\,0.010$ \\
  \bottomrule
\end{tabular}
}
\end{table}

\subsection{TabPFN v2.5 comparison}
\label{app:tabpfn_v25}

The TabPFN v2.5 comparison repeats PFN-NPE with the updated TabPFN encoder on
tasks with paired seeds. The three tables report joint, marginal, and rank
C2ST. The \(\Delta\) column is v2.5 minus v2, paired by seed, so negative
values mean lower C2ST with v2.5 on matched runs.

\begin{table}[t]
  \centering
  \caption{TabPFN v2.5 joint C2ST comparison.}
  \label{tab:tabpfn_v25_joint}
  % Auto-generated by scripts/tabpfn_v25_comparison_table.py. Do not edit by hand.
\begin{tabular}{lcccc}
  \toprule
  Task & Seeds & TabPFN v2 & TabPFN v2.5 & $\Delta$ \\
  \midrule
  \texttt{slcp} & 5 & $0.826\,\pm\,0.01$ & $\mathbf{0.799}\,\pm\,0.01$ & $-0.027\,\pm\,0.02$ \\
  \texttt{slcp\_distractors} & 5 & $0.861\,\pm\,0.01$ & $\mathbf{0.831}\,\pm\,0.01$ & $-0.030\,\pm\,0.02$ \\
  \texttt{bernoulli\_glm\_distractors} & 5 & $\mathbf{0.673}\,\pm\,0.01$ & $0.682\,\pm\,0.01$ & $+0.009\,\pm\,0.02$ \\
  \texttt{two\_moons\_distractors} & 5 & $0.597\,\pm\,0.02$ & $\mathbf{0.576}\,\pm\,0.02$ & $-0.021\,\pm\,0.01$ \\
  \texttt{ou} & 3 & $\mathbf{0.866}\,\pm\,0.01$ & $0.892\,\pm\,0.00$ & $+0.026\,\pm\,0.01$ \\
  \bottomrule
\end{tabular}

\end{table}

\begin{table}[t]
  \centering
  \caption{TabPFN v2.5 marginal C2ST comparison.}
  \label{tab:tabpfn_v25_marginal}
  % Auto-generated by scripts/tabpfn_v25_comparison_table.py. Do not edit by hand.
\begin{tabular}{lcccc}
  \toprule
  Task & Seeds & TabPFN v2 & TabPFN v2.5 & $\Delta$ \\
  \midrule
  \texttt{slcp} & 5 & $0.620\,\pm\,0.01$ & $\mathbf{0.606}\,\pm\,0.01$ & $-0.013\,\pm\,0.01$ \\
  \texttt{slcp\_distractors} & 5 & $0.640\,\pm\,0.01$ & $\mathbf{0.621}\,\pm\,0.01$ & $-0.019\,\pm\,0.01$ \\
  \texttt{bernoulli\_glm\_distractors} & 5 & $\mathbf{0.562}\,\pm\,0.00$ & $0.564\,\pm\,0.01$ & $+0.001\,\pm\,0.01$ \\
  \texttt{two\_moons\_distractors} & 5 & $0.528\,\pm\,0.01$ & $\mathbf{0.526}\,\pm\,0.01$ & $-0.002\,\pm\,0.01$ \\
  \texttt{ou} & 3 & $\mathbf{0.722}\,\pm\,0.00$ & $0.748\,\pm\,0.01$ & $+0.025\,\pm\,0.01$ \\
  \bottomrule
\end{tabular}

\end{table}

\begin{table}[t]
  \centering
  \caption{TabPFN v2.5 rank C2ST comparison.}
  \label{tab:tabpfn_v25_rank}
  % Auto-generated by scripts/tabpfn_v25_comparison_table.py. Do not edit by hand.
\begin{tabular}{lcccc}
  \toprule
  Task & Seeds & TabPFN v2 & TabPFN v2.5 & $\Delta$ \\
  \midrule
  \texttt{slcp} & 5 & $0.833\,\pm\,0.01$ & $\mathbf{0.812}\,\pm\,0.01$ & $-0.021\,\pm\,0.01$ \\
  \texttt{slcp\_distractors} & 5 & $0.863\,\pm\,0.01$ & $\mathbf{0.836}\,\pm\,0.02$ & $-0.028\,\pm\,0.02$ \\
  \texttt{bernoulli\_glm\_distractors} & 5 & $\mathbf{0.670}\,\pm\,0.01$ & $0.672\,\pm\,0.01$ & $+0.002\,\pm\,0.02$ \\
  \texttt{two\_moons\_distractors} & 5 & $0.640\,\pm\,0.02$ & $\mathbf{0.610}\,\pm\,0.02$ & $-0.029\,\pm\,0.02$ \\
  \texttt{ou} & 3 & $\mathbf{0.880}\,\pm\,0.01$ & $0.903\,\pm\,0.00$ & $+0.024\,\pm\,0.02$ \\
  \bottomrule
\end{tabular}

\end{table}

\subsection{Projection-capacity ablation}
\label{app:capacity_ablation}

The default PFN-NPE configuration compresses the concatenated per-parameter
TabPFN embedding to a 64-dimensional summary before fitting the NSF posterior
estimator. We test whether this projection limits joint posterior recovery on
SLCP by repeating the experiment with a wider learned linear projector and with
the raw concatenated embedding. All three settings use the same training
budget, validation split, random seeds, reference observations, and NSF
training protocol.

\subsection{PCA projection ablation}
\label{app:pca_no_pca}

The PCA ablation tests whether the default PCA-64 projection limits the final
posterior estimator. The first table compares the default projection with full
embeddings across the broad \(10^4\)-simulation set and a smaller
\(5{\times}10^4\)-simulation core set. The \(\Delta\) column is no-PCA minus
PCA-64. The second table focuses on hard tasks and compares the default PCA-64
NSF estimator, the standard NSF estimator on full embeddings, and an XL NSF
estimator on full embeddings with common seeds.

\begin{table}[t]
  \centering
  \caption{PCA-64 versus no-PCA PFN-NPE.}
  \label{tab:pca_no_pca_budget}
  \resizebox{\linewidth}{!}{% Auto-generated by scripts/pca_no_pca_table.py. Do not edit by hand.
\begin{tabular}{llccc}
  \toprule
  Budget & Task & PCA-64 & No PCA & $\Delta$ \\
  \midrule
  $10^4$ & Two moons & $0.584\,\pm\,0.01$ & $\mathbf{0.583}\,\pm\,0.01$ & -0.001 \\
  $10^4$ & Gaussian mixture & $\mathbf{0.607}\,\pm\,0.05$ & $0.608\,\pm\,0.05$ & +0.001 \\
  $10^4$ & Gaussian linear & $\mathbf{0.552}\,\pm\,0.01$ & $0.553\,\pm\,0.01$ & +0.001 \\
  $10^4$ & Gaussian linear uniform & $\mathbf{0.588}\,\pm\,0.00$ & $0.590\,\pm\,0.01$ & +0.002 \\
  $10^4$ & Bernoulli GLM & $\mathbf{0.658}\,\pm\,0.00$ & $0.660\,\pm\,0.00$ & +0.002 \\
  $10^4$ & SLCP & $0.829\,\pm\,0.02$ & $\mathbf{0.828}\,\pm\,0.02$ & +0.000 \\
  $10^4$ & SIR & $\mathbf{0.611}\,\pm\,0.03$ & $0.612\,\pm\,0.03$ & +0.001 \\
  $10^4$ & Lotka-Volterra & $\mathbf{0.931}\,\pm\,0.01$ & $0.932\,\pm\,0.01$ & +0.001 \\
  $10^4$ & G. mixture + distractors & $\mathbf{0.626}\,\pm\,0.04$ & $0.655\,\pm\,0.04$ & +0.029 \\
  $10^4$ & Two moons + distractors & $0.576\,\pm\,0.02$ & $\mathbf{0.568}\,\pm\,0.01$ & -0.008 \\
  $10^4$ & Bernoulli GLM + distractors & $\mathbf{0.685}\,\pm\,0.01$ & $0.686\,\pm\,0.00$ & +0.001 \\
  $10^4$ & SIR + distractors & $\mathbf{0.620}\,\pm\,0.01$ & $0.623\,\pm\,0.01$ & +0.004 \\
  $10^4$ & AR(1), T=50 & $\mathbf{0.783}\,\pm\,0.02$ & $0.783\,\pm\,0.02$ & +0.000 \\
  $10^4$ & OU & $0.866\,\pm\,0.01$ & $\mathbf{0.861}\,\pm\,0.01$ & -0.005 \\
  $10^4$ & Solar dynamo & $\mathbf{0.868}\,\pm\,0.01$ & $0.879\,\pm\,0.00$ & +0.011 \\
  \midrule
  $5{\times}10^4$ & Two moons & $0.534\,\pm\,0.00$ & $\mathbf{0.533}\,\pm\,0.00$ & -0.001 \\
  $5{\times}10^4$ & Bernoulli GLM & $\mathbf{0.606}\,\pm\,0.01$ & $0.607\,\pm\,0.01$ & +0.001 \\
  $5{\times}10^4$ & SLCP & $0.743\,\pm\,0.02$ & $\mathbf{0.742}\,\pm\,0.02$ & -0.001 \\
  $5{\times}10^4$ & G. mixture + distractors & $0.562\,\pm\,0.02$ & $\mathbf{0.553}\,\pm\,0.02$ & -0.008 \\
  $5{\times}10^4$ & AR(1), T=50 & $\mathbf{0.771}\,\pm\,0.01$ & $0.771\,\pm\,0.01$ & +0.000 \\
  \bottomrule
\end{tabular}
}
\end{table}

\begin{table}[t]
  \centering
  \caption{XL no-PCA flow-estimator diagnostic.}
  \label{tab:xl_no_pca}
  \resizebox{\linewidth}{!}{% Auto-generated by scripts/xl_no_pca_table.py. Do not edit by hand.
\begin{tabular}{llcccc}
  \toprule
  Budget & Task & PCA-64 NSF & No-PCA NSF & XL no-PCA NSF & $\Delta_{\mathrm{XL-PCA}}$ \\
  \midrule
  $10^4$ & SLCP & $0.829\,\pm\,0.02$ & $0.828\,\pm\,0.02$ & $\mathbf{0.822}\,\pm\,0.01$ & -0.006 \\
  $10^4$ & Lotka-Volterra & $\mathbf{0.931}\,\pm\,0.01$ & $0.932\,\pm\,0.01$ & $0.943\,\pm\,0.00$ & +0.012 \\
  $10^4$ & OU & $0.866\,\pm\,0.01$ & $0.861\,\pm\,0.01$ & $\mathbf{0.856}\,\pm\,0.00$ & -0.010 \\
  $10^4$ & AR(1), T=50 & $\mathbf{0.783}\,\pm\,0.02$ & $0.783\,\pm\,0.02$ & $0.785\,\pm\,0.01$ & +0.002 \\
  $10^4$ & Solar dynamo & $\mathbf{0.868}\,\pm\,0.01$ & $0.879\,\pm\,0.00$ & $0.888\,\pm\,0.01$ & +0.020 \\
  \midrule
  $5{\times}10^4$ & SLCP & $0.743\,\pm\,0.02$ & $\mathbf{0.742}\,\pm\,0.02$ & $0.751\,\pm\,0.02$ & +0.008 \\
  $5{\times}10^4$ & Lotka-Volterra & $\mathbf{0.906}\,\pm\,0.02$ & --- & $0.920\,\pm\,0.01$ & +0.014 \\
  $5{\times}10^4$ & OU & $0.857\,\pm\,0.01$ & --- & $\mathbf{0.856}\,\pm\,0.01$ & -0.001 \\
  $5{\times}10^4$ & AR(1), T=50 & $\mathbf{0.771}\,\pm\,0.01$ & $0.771\,\pm\,0.01$ & $0.773\,\pm\,0.01$ & +0.002 \\
  $5{\times}10^4$ & Solar dynamo & $\mathbf{0.863}\,\pm\,0.01$ & --- & $0.873\,\pm\,0.01$ & +0.010 \\
  \bottomrule
\end{tabular}
}
\end{table}

\subsection{Parameter-role probe summary}
\label{app:probe_parameter_roles}

The parameter-role table summarizes final-layer cross-\(\theta\) probes for
task parameters with interpretable simulator roles. Matched probes decode
\(\theta_j\) from the matching target chunk \(e_j\), and the off-mean column
averages nonmatching chunks. The \(\Delta\) column is matched minus off mean.
The table is descriptive because parameter semantics and identifiability differ
across simulators.

\begin{table}[t]
  \centering
  \caption{Final-layer parameter-role probe summary.}
  \label{tab:probe_parameter_roles}
  \resizebox{\linewidth}{!}{% Auto-generated by scripts/plot_probe_main_and_role_table.py. Do not edit by hand.
\begin{tabular}{lllrrr}
  \toprule
  Task & Parameter & Role & Matched & Off mean & $\Delta$ \\
  \midrule
  SLCP & $\theta_0$ mean 1 & location & 0.60 $\pm$ 0.02 & 0.41 $\pm$ 0.02 & 0.20 $\pm$ 0.03 \\
  SLCP & $\theta_1$ mean 2 & location & 0.60 $\pm$ 0.02 & 0.43 $\pm$ 0.03 & 0.18 $\pm$ 0.04 \\
  SLCP & $\theta_2$ scale 1 & scale & -0.00 $\pm$ 0.00 & -0.00 $\pm$ 0.00 & -0.00 $\pm$ 0.00 \\
  SLCP & $\theta_3$ scale 2 & scale & -0.00 $\pm$ 0.00 & -0.00 $\pm$ 0.00 & -0.00 $\pm$ 0.00 \\
  SLCP & $\theta_4$ corr. & correlation & 0.38 $\pm$ 0.02 & 0.01 $\pm$ 0.00 & 0.37 $\pm$ 0.02 \\
  \midrule
  SLCP + distr. & $\theta_0$ mean 1 & location & 0.60 $\pm$ 0.02 & 0.19 $\pm$ 0.02 & 0.41 $\pm$ 0.03 \\
  SLCP + distr. & $\theta_1$ mean 2 & location & 0.60 $\pm$ 0.02 & 0.20 $\pm$ 0.03 & 0.40 $\pm$ 0.05 \\
  SLCP + distr. & $\theta_2$ scale 1 & scale & -0.01 $\pm$ 0.01 & -0.00 $\pm$ 0.00 & -0.01 $\pm$ 0.01 \\
  SLCP + distr. & $\theta_3$ scale 2 & scale & -0.00 $\pm$ 0.00 & -0.00 $\pm$ 0.00 & -0.00 $\pm$ 0.00 \\
  SLCP + distr. & $\theta_4$ corr. & correlation & 0.00 $\pm$ 0.00 & -0.00 $\pm$ 0.00 & 0.00 $\pm$ 0.00 \\
  \midrule
  OU & $\alpha$ long-run mean & location & 0.97 $\pm$ 0.00 & 0.91 $\pm$ 0.01 & 0.06 $\pm$ 0.01 \\
  OU & $\beta$ reversion rate & rate & 0.59 $\pm$ 0.01 & 0.32 $\pm$ 0.01 & 0.27 $\pm$ 0.02 \\
  OU & $\sigma$ diffusion scale & scale & 0.93 $\pm$ 0.00 & 0.86 $\pm$ 0.00 & 0.07 $\pm$ 0.00 \\
  \midrule
  Solar dynamo & $\alpha_{\min}$ growth floor & location/rate & 0.98 $\pm$ 0.00 & 0.96 $\pm$ 0.00 & 0.02 $\pm$ 0.00 \\
  Solar dynamo & $\alpha_{\mathrm{range}}$ growth range & scale/range & 0.69 $\pm$ 0.01 & 0.67 $\pm$ 0.01 & 0.02 $\pm$ 0.01 \\
  Solar dynamo & $\epsilon_{\max}$ noise bound & scale & 0.14 $\pm$ 0.03 & 0.17 $\pm$ 0.01 & -0.04 $\pm$ 0.01 \\
  \midrule
  AR(1) & $\rho$ autocorrelation & correlation/rate & 0.82 $\pm$ 0.01 & 0.19 $\pm$ 0.00 & 0.63 $\pm$ 0.01 \\
  AR(1) & $\log\sigma$ innovation scale & scale & 0.96 $\pm$ 0.00 & 0.96 $\pm$ 0.00 & -0.00 $\pm$ 0.00 \\
  \midrule
  SIR & $\beta$ infection rate & rate & 0.98 $\pm$ 0.01 & 0.91 $\pm$ 0.02 & 0.07 $\pm$ 0.02 \\
  SIR & $\gamma$ recovery rate & rate & 0.83 $\pm$ 0.01 & 0.79 $\pm$ 0.02 & 0.04 $\pm$ 0.01 \\
  \midrule
  Lotka-Volterra & $\alpha$ prey growth & rate & 0.98 $\pm$ 0.01 & 0.82 $\pm$ 0.01 & 0.16 $\pm$ 0.02 \\
  Lotka-Volterra & $\beta$ predation & interaction & 0.91 $\pm$ 0.01 & 0.71 $\pm$ 0.00 & 0.20 $\pm$ 0.00 \\
  Lotka-Volterra & $\gamma$ predator death & rate & 0.96 $\pm$ 0.01 & 0.82 $\pm$ 0.01 & 0.13 $\pm$ 0.00 \\
  Lotka-Volterra & $\delta$ predator growth & interaction & 0.95 $\pm$ 0.01 & 0.77 $\pm$ 0.01 & 0.18 $\pm$ 0.01 \\
  \bottomrule
\end{tabular}
}
\end{table}

\subsection{Raw-observation quantile-probe ablation}
\label{app:quantile_raw_ablation}

The raw-observation ablation compares quantile probes trained on frozen PFN
summaries with probes trained on standardized raw observations \(x\). Each row
uses the same task, seeds, training and validation sizes, ten reference
observations, and quantile levels
\(\tau \in \{0.05,0.25,0.5,0.75,0.95\}\). Correlations measure
predicted-versus-reference marginal quantile variation. Pinball ratios divide
probe pinball loss by empirical-reference-quantile pinball loss on the same
reference observations; \(1\times\) is the empirical reference target. Raw-task
variants are excluded to match the paper-facing benchmark set.

\begin{table}[t]
  \centering
  \caption{Raw-observation quantile-probe ablation.}
  \label{tab:quantile_raw_ablation}
  \resizebox{\linewidth}{!}{% Auto-generated by scripts/quantile_raw_ablation_table.py. Do not edit by hand.
\begin{tabular}{lrrrrr}
  \toprule
  Task & $n$ & PFN $r$ & Raw $x$ $r$ & PFN pinball ratio & Raw $x$ pinball ratio \\
  \midrule
  Two moons & 3 & 0.973 $\pm$ 0.010 & 0.985 $\pm$ 0.001 & 1.14 $\pm$ 0.07 & 1.04 $\pm$ 0.02 \\
  Gaussian mixture & 3 & 1.000 $\pm$ 0.000 & 1.000 $\pm$ 0.000 & 1.01 $\pm$ 0.00 & 1.01 $\pm$ 0.00 \\
  Gaussian linear & 3 & 0.994 $\pm$ 0.000 & 0.998 $\pm$ 0.000 & 1.01 $\pm$ 0.01 & 1.00 $\pm$ 0.00 \\
  Gaussian linear uniform & 3 & 0.999 $\pm$ 0.000 & 0.984 $\pm$ 0.000 & 1.01 $\pm$ 0.01 & 1.09 $\pm$ 0.00 \\
  Bernoulli GLM & 3 & 0.995 $\pm$ 0.002 & 0.971 $\pm$ 0.000 & 1.15 $\pm$ 0.01 & 1.89 $\pm$ 0.00 \\
  SIR & 3 & 0.992 $\pm$ 0.001 & 0.972 $\pm$ 0.001 & 2.43 $\pm$ 0.27 & 5.28 $\pm$ 0.06 \\
  Lotka-Volterra & 3 & 0.992 $\pm$ 0.001 & 0.875 $\pm$ 0.005 & 2.71 $\pm$ 0.09 & 11.05 $\pm$ 0.11 \\
  SLCP & 3 & 0.846 $\pm$ 0.025 & 0.470 $\pm$ 0.005 & 1.32 $\pm$ 0.06 & 1.78 $\pm$ 0.04 \\
  SLCP + distractors & 3 & 0.676 $\pm$ 0.009 & 0.460 $\pm$ 0.009 & 1.49 $\pm$ 0.02 & 1.79 $\pm$ 0.01 \\
  Two moons + distractors & 3 & 0.967 $\pm$ 0.023 & 0.985 $\pm$ 0.003 & 1.11 $\pm$ 0.03 & 1.05 $\pm$ 0.01 \\
  Gaussian mixture + distractors & 3 & 1.000 $\pm$ 0.000 & 1.000 $\pm$ 0.000 & 0.99 $\pm$ 0.10 & 1.04 $\pm$ 0.01 \\
  Bernoulli GLM + distractors & 3 & 0.990 $\pm$ 0.001 & 0.970 $\pm$ 0.000 & 1.28 $\pm$ 0.04 & 1.90 $\pm$ 0.01 \\
  SIR + distractors & 3 & 0.997 $\pm$ 0.000 & 0.970 $\pm$ 0.001 & 3.01 $\pm$ 0.31 & 5.32 $\pm$ 0.07 \\
  AR(1) time series & 3 & 0.974 $\pm$ 0.006 & 0.585 $\pm$ 0.014 & 1.88 $\pm$ 0.18 & 5.53 $\pm$ 0.05 \\
  Ornstein-Uhlenbeck & 3 & 0.921 $\pm$ 0.043 & 0.949 $\pm$ 0.000 & 4.64 $\pm$ 1.21 & 2.65 $\pm$ 0.03 \\
  Solar dynamo & 3 & 0.998 $\pm$ 0.001 & 0.996 $\pm$ 0.000 & 2.27 $\pm$ 0.05 & 3.04 $\pm$ 0.08 \\
  \bottomrule
\end{tabular}
}
\end{table}

\clearpage
\section{Additional Case Studies}
\label{app:case_studies}

This appendix expands two case studies used to inspect posterior errors beyond
aggregate C2ST. Each case study combines a parameter-specific cross-\(\theta\)
probe with posterior-shape diagnostics. The probe splits the concatenated
TabPFN summary into parameter-indexed chunks \(e_j\) and fits ridge regressors
that decode \(\theta_j\) from the matching chunk \(e_j\) and from nonmatching
chunks \(e_i\), \(i\neq j\). The posterior diagnostics use reference
observation 1 for marginal KDE and bivariate contour displays. The moment and
highest-density-region (HDR) summaries aggregate across available reference
observations and seeds. Mean errors are measured in reference-posterior
standard-deviation units. Dispersion and HDR quantities are reported as log
ratios relative to the reference posterior. Values near zero mean closer
agreement with the reference summaries.

\subsection{SLCP case study}
\label{app:slcp_probe_case_study}
\label{app:slcp_failure_diagnostics}

\begin{figure}[p]
  \centering
  \includegraphics[width=\linewidth,height=0.82\textheight,keepaspectratio]{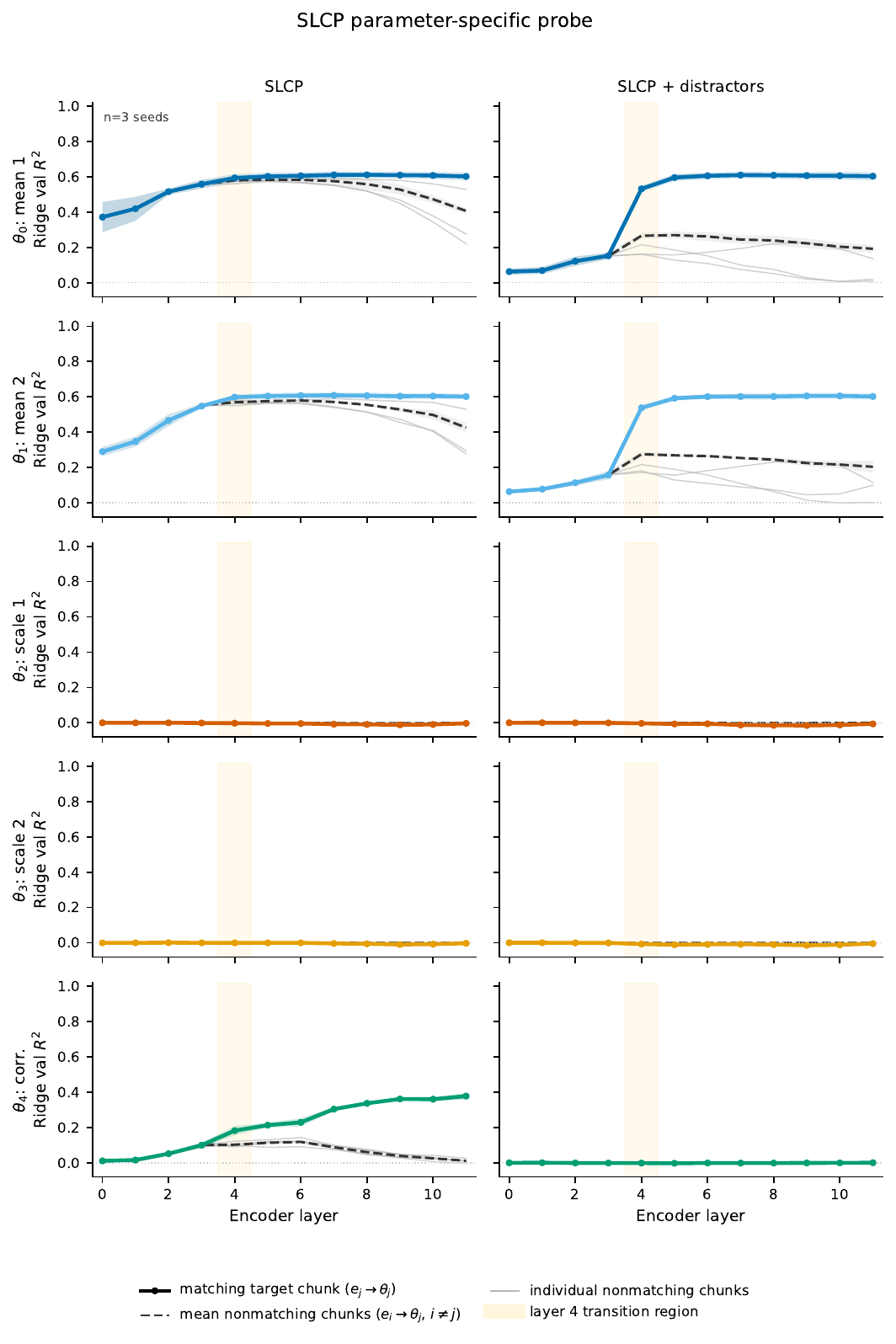}
  \caption{SLCP parameter-specific cross-$\theta$ probe.}
  \label{fig:slcp_probe_target_accessibility}
\end{figure}

The SLCP probe in \cref{fig:slcp_probe_target_accessibility} shows
heterogeneous parameter accessibility in the frozen summaries. The two location
coordinates are decodable from their matching chunks in both the original and
distractor variants, with final-layer validation \(R^2\) values around 0.60.
The nonmatching-chunk average for these location parameters is about
0.41--0.43 in the original task and about 0.19--0.20 after adding distractors.
The two scale coordinates have validation \(R^2\) values near zero for both
matching and nonmatching chunks. The correlation-like coordinate has
matched-chunk \(R^2 = 0.38\) in the original task and about 0.00 in the
distractor variant. These results suggest that the summaries preserve location
information, carry little linearly accessible scale information under this
probe, and encode a correlation-like signal that is sensitive to observation
augmentation.

\begin{figure}[p]
  \centering
  \includegraphics[width=\linewidth]{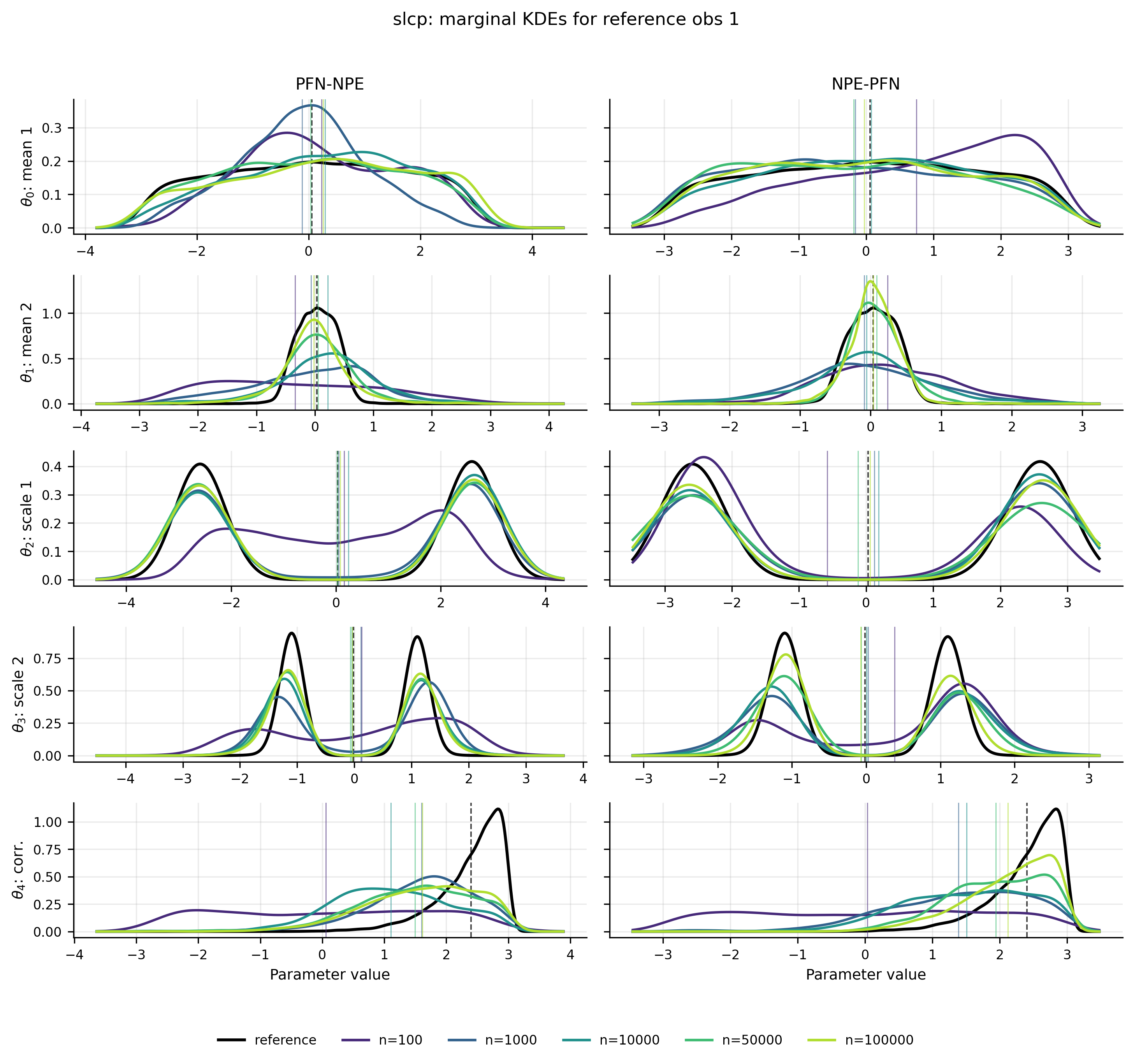}
  \caption{SLCP marginal posterior KDEs across simulation budgets.}
  \label{fig:slcp_failure_kde_budget_overlay}
\end{figure}

\begin{figure}[p]
  \centering
  \includegraphics[width=\linewidth,height=0.78\textheight,keepaspectratio]{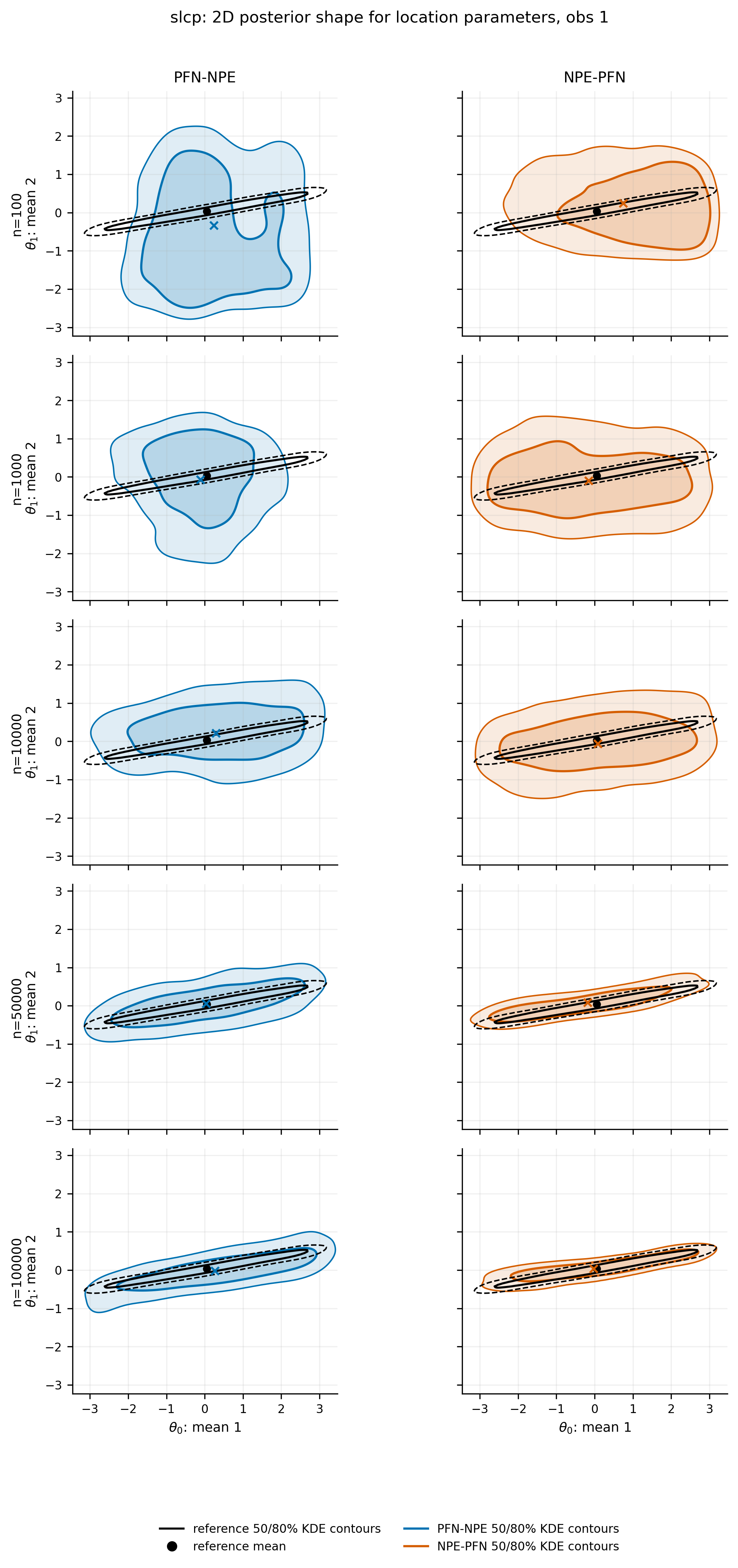}
  \caption{SLCP bivariate location-posterior KDEs across simulation budgets.}
  \label{fig:slcp_failure_location2d_kde}
\end{figure}

\begin{figure}[p]
  \centering
  \includegraphics[width=\linewidth,height=0.82\textheight,keepaspectratio]{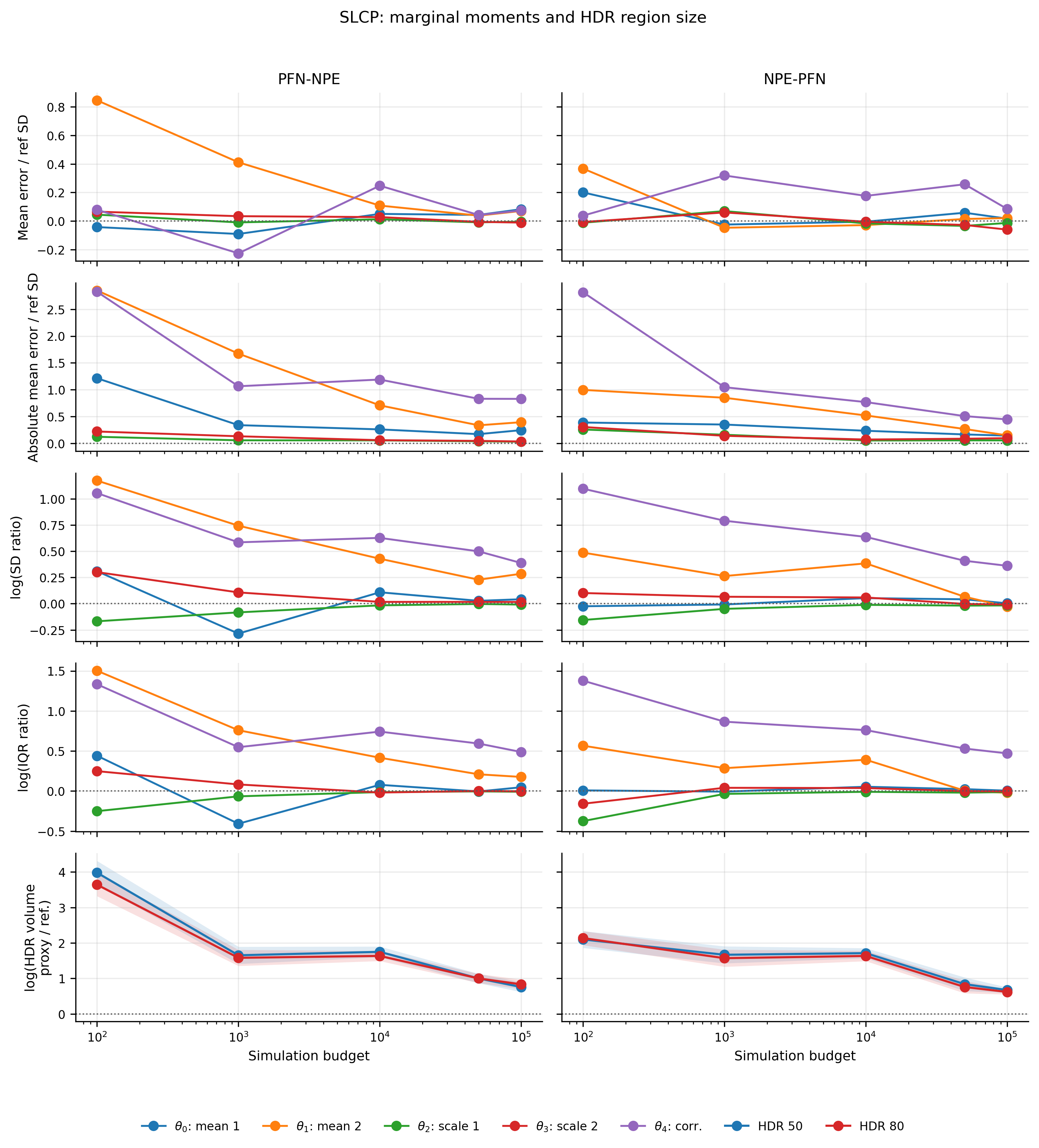}
  \caption{SLCP marginal moment and HDR region-size diagnostics.}
  \label{fig:slcp_failure_moment_trajectories}
  \label{fig:slcp_failure_hdr}
\end{figure}

The posterior-shape diagnostics in
\cref{fig:slcp_failure_kde_budget_overlay,fig:slcp_failure_location2d_kde,fig:slcp_failure_moment_trajectories}
show how these representation patterns appear in posterior samples. The
marginal KDEs compare PFN-NPE and NPE-PFN against the SBIBM reference
posterior, with colored curves for increasing simulation budgets and vertical
lines for marginal posterior means. The location marginals and
bivariate location contours move toward the reference geometry as the budget
grows. The scale and correlation-like marginals retain visible shape mismatch,
especially at low budgets. Across reference observations and seeds,
the median absolute marginal mean error falls from 0.65 to 0.10 for PFN-NPE and
from 0.43 to 0.10 for NPE-PFN over the reported budgets. At high budgets, most
marginal mean and dispersion curves lie close to zero, while the HDR region-size
proxies remain positive, indicating residual joint-region inflation.

\subsection{AR(1) case study}
\label{app:ar1_probe_case_study}
\label{app:ar1_failure_diagnostics}

\begin{figure}[p]
  \centering
  \includegraphics[width=0.72\linewidth]{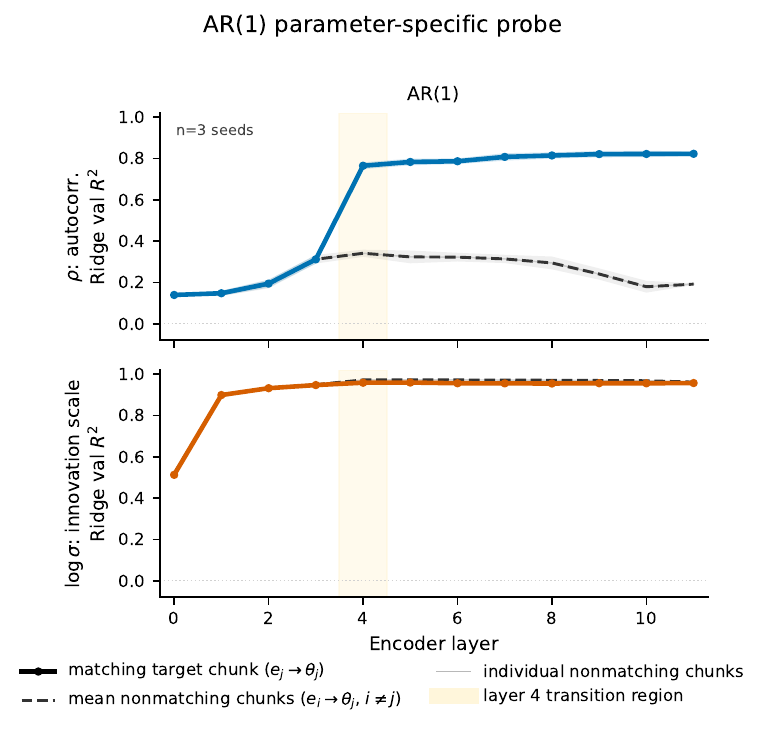}
  \caption{AR(1) parameter-specific cross-$\theta$ probe.}
  \label{fig:ar1_probe_target_accessibility}
\end{figure}

The AR(1) probe in \cref{fig:ar1_probe_target_accessibility} gives a
two-parameter view of the same diagnostic. The autocorrelation parameter
\(\rho\) has final-layer validation \(R^2 = 0.82\) from its matching chunk and
\(R^2 = 0.19\) from the nonmatching chunk. The innovation-scale parameter
\(\log\sigma\) has validation \(R^2\) values of about 0.96 from both chunks.
In this task, autocorrelation information is concentrated in its
coordinate-specific chunk, while innovation-scale information is broadly
available across chunks.

The AR(1) posterior diagnostics focus on \((\rho,\log\sigma)\) for reference
observation 1. The marginal KDEs and bivariate contours compare PFN-NPE and
NPE-PFN against the grid-reference posterior, using the same conventions as the
SLCP diagnostics. Low-budget posterior samples show large errors in
\(\rho\), and increasing the budget moves both marginal and bivariate summaries
toward the reference posterior. The moment and HDR trajectories show the same
budget trend across reference observations. Mean error for \(\rho\) contracts
sharply from \(n = 100\) to \(n = 10{,}000\) for both displayed methods. The
dispersion and HDR log ratios remain positive at high budget, so the remaining
error involves posterior width and joint region size in addition to marginal
location.

\begin{figure}[p]
  \centering
  \includegraphics[width=\linewidth]{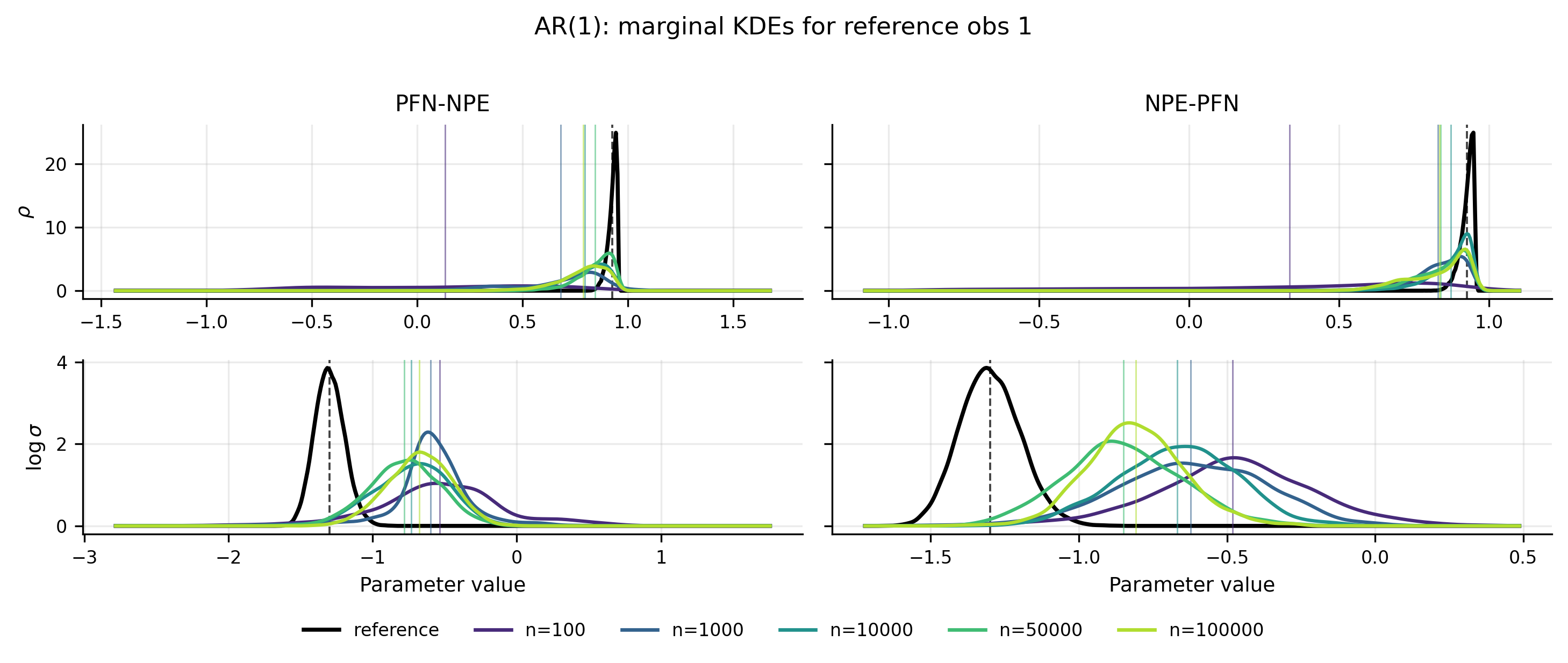}
  \caption{AR(1) marginal posterior KDEs across simulation budgets.}
  \label{fig:ar1_failure_kde_budget_overlay}
\end{figure}

\begin{figure}[p]
  \centering
  \includegraphics[width=\linewidth,height=0.78\textheight,keepaspectratio]{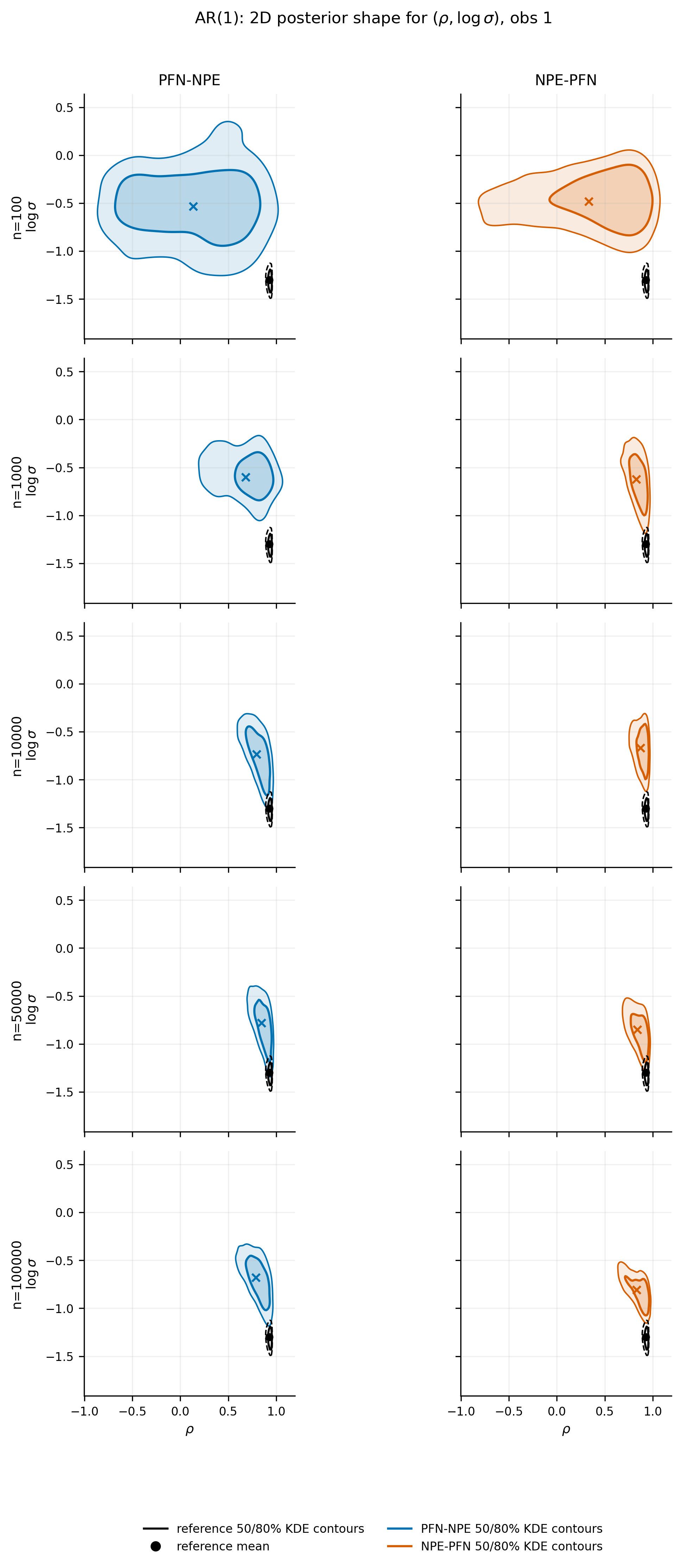}
  \caption{AR(1) bivariate posterior KDEs across simulation budgets.}
  \label{fig:ar1_failure_location2d_kde}
\end{figure}

\begin{figure}[p]
  \centering
  \includegraphics[width=\linewidth,height=0.82\textheight,keepaspectratio]{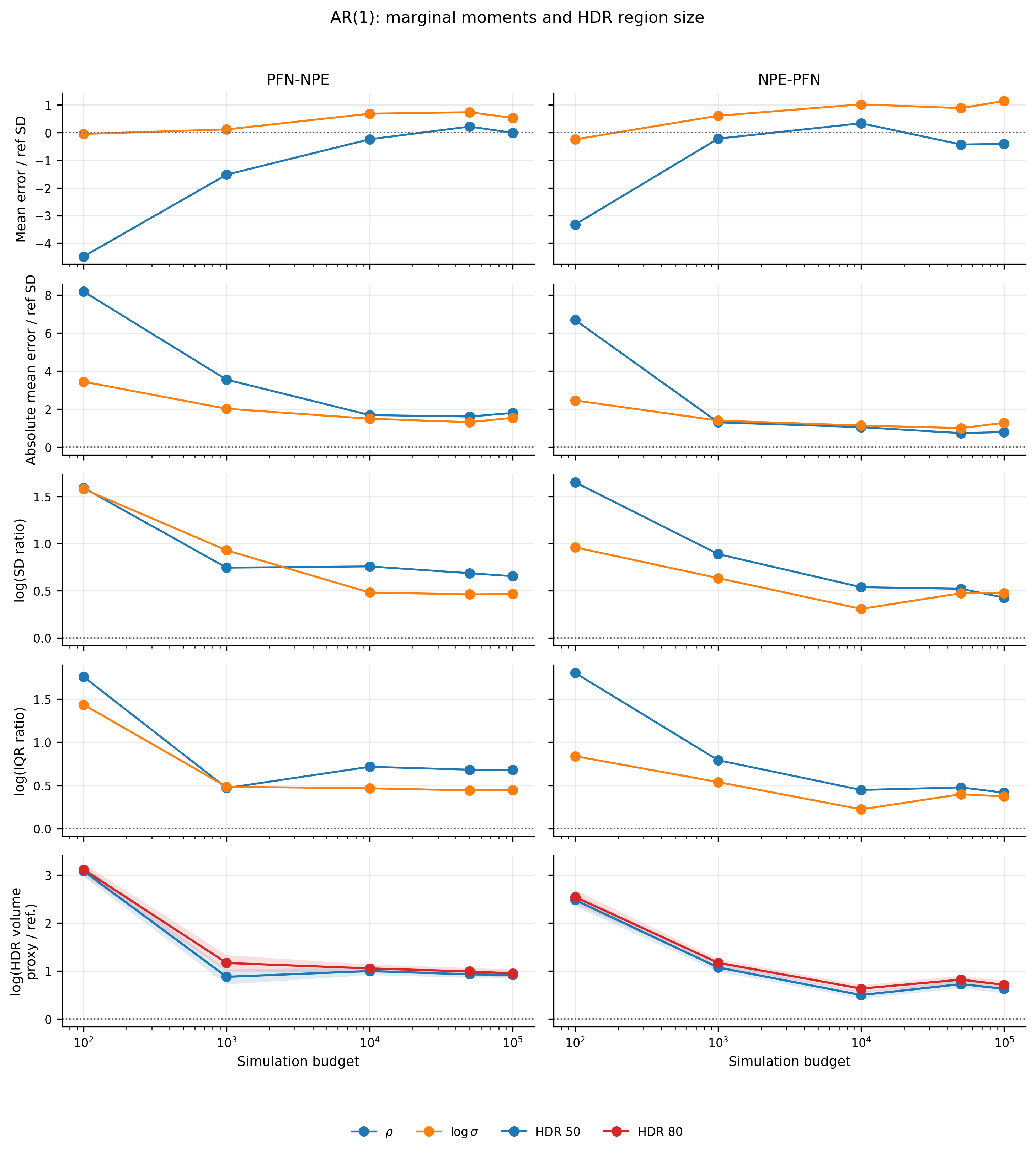}
  \caption{AR(1) marginal moment and HDR region-size diagnostics.}
  \label{fig:ar1_failure_moment_trajectories}
  \label{fig:ar1_failure_hdr}
\end{figure}

\clearpage

% \clearpage
% \input{checklist}

\end{document}